\documentclass[natbib]{svmult}

\usepackage{proof}
\usepackage{amsmath}
\usepackage{tikz}
\usetikzlibrary{arrows,matrix,graphs,shapes,snakes,automata,backgrounds,petri,positioning,calc,decorations.pathmorphing}
\usepackage{amssymb}
\usepackage{natbib}
\usepackage{linguex}
\usepackage{marvosym}
\usepackage{wrapfig}
\usepackage{mathptmx}
\usepackage{helvet}
\usepackage{courier}
\usepackage{makeidx}
\usepackage{graphicx}
\usepackage{multicol}
\usepackage{footmisc}

\newcommand{\chapmeryretore}{6}
\newcommand{\citeasnoun}[1]{\citet{#1}}
\newcommand{\bo}{[}
\newcommand{\bc}{]}
\newcommand{\bs}{\backslash}
\newcommand{\ra}{\vdash}
\newcommand{\Gammap}{\Gamma^{\prime}}
\newcommand{\p}[3]{#2 \circ_{#1} #3}

\newcommand{\calr}{\mathcal{R}}
\newcommand{\lolli}{\multimap}
\newcommand{\editout}[1]{}

\colorlet{lightgray}{gray!10}

\newcommand{\apsnodei}{\centerdot_{\rule{0pt}{1.2ex}}}
\newcommand{\ldr}{\mathbin{/}}
\newcommand{\ldl}{\mathbin{\backslash}}
\newcommand{\lpr}{\bullet}

\tikzstyle{reverseclip}=[insert path={(current page.north east) --
  (current page.south east) --
  (current page.south west) --
  (current page.north west) --
  (current page.north east)}
]

\tikzset{pas/.style={fill=gray!60}, 
act/.style={fill=gray!30},
main/.style={draw,fill=white},
ctx/.style={rounded rectangle,minimum size=7mm},
val/.style={rectangle,minimum size=7mm},
cmd/.style={chamfered rectangle,draw,fill=white},
tns/.style={circle,minimum size=4mm,draw,fill=white},
par/.style={circle,minimum size=4mm,draw,fill=black}, 
minipar/.style={circle,minimum size=2.5mm,draw,fill=black}, 
pn/.style={rounded corners, rectangle,fill=blue!30,draw,minimum size=15mm},
medpn/.style={rounded corners, rectangle,fill=blue!30,draw,minimum size=20mm},
 bigpn/.style={rounded corners, rectangle,fill=blue!30,draw,minimum size=25mm}}

\title{The Grail theorem prover: \\Type theory for syntax and semantics}
\author{Richard Moot}
\institute{Richard Moot \at CNRS, LaBRI, Bordeaux University, 351 cours de la Lib\'{e}ration, 33405 Talence, France \and LIRMM, Montpellier University, 161 rue Ada,
34095 Montpellier cedex 5, France \email{Richard.Moot@labri.fr}}

\begin{document}
\maketitle

%\begin{abstract}
\abstract{Type-logical grammars use a foundation of logic and type theory to model natural language. These grammars have
been particularly successful giving an account of several well-known phenomena on the syntax-semantics interface, such as quantifier scope and its interaction with other phenomena. This chapter gives a high-level description of a family of theorem provers
designed for grammar development in a variety of modern type-logical grammars. We discuss automated theorem proving for type-logical grammars from the perspective
of proof nets, a graph-theoretic way to represent (partial) proofs during proof search.}
%\end{abstract}

\section{Introduction}

This chapter describes a series of tools for developing and
testing type-logical grammars. The Grail family of theorem provers
have been designed to work with a variety of modern type-logical
frameworks, including multimodal type-logical grammars \citep{Moo11},
NL$_{cl}$ \citep{bs14cont}, the
Displacement calculus \citep{mvf11displacement} and hybrid type-logical
grammars \citep{kl12gap}.

The tools give a transparent way of implementing grammars and testing
their consequences, providing a natural deduction proof in the
specific type-logical grammar for each of the readings of a sentence. None of this
replaces careful reflection by the grammar writer, of course, but in many cases, computational testing of
hand-written grammars will reveal surprises, showing unintended
consequences of our grammar and such unintended proofs (or unintended
\emph{absences} of proofs) help us improve the grammar. Computational
tools also help us speed up grammar development, for example by
allowing us to compare
several alternative solutions to a problem and investigate where they make different predictions.
%Much like we can never truly prove something is safe, we cannot show a
%given grammar does not overgenerate since it depends on asking the
%right grammaticality questions (though, following Sapir, ``all
%grammars leak'' not all such leaks are equally serious).

This chapter describes the underlying formalism of the theorem
provers, as it is visible during an interactive proof trace, and
present the general strategy followed by the theorem provers. The
presentation in this chapter is somewhat informal, referring the
reader elsewhere for full proofs.

The rest of this chapter is structured as follows. Section~\ref{sec:tlg} presents a general introduction to
type-logical grammars and illustrates its basic concepts using the
Lambek calculus, ending the section with some problems at the
syntax-semantics interface for the Lambek
calculus. Section~\ref{sec:mtlg} looks at recent developments in
type-logical grammars and how they solve some of the problems at the
syntax-semantics interface. Section~\ref{sec:parse} looks at two
general frameworks for
automated theorem proving for type-logical grammars, describing the
internal representation of partial proofs and giving a high-level
overview of the proof search mechanism.

\section{Type-logical grammars}
\label{sec:tlg}

Type-logical grammars are a family of grammar formalisms built on a foundation of logic and type theory. Type-logical grammars originated when \citeasnoun{lambek} introduced his Syntactic Calculus (called the Lambek calculus, L, by later authors). Though Lambek built on the work of \citeasnoun{ajd}, \citeasnoun{quasi} and others, Lambek's main innovation was to cast the calculus as a logic, giving a sequent calculus and showing decidability by means of cut elimination. This combination of linguistic and computational applications has proved very influential.

In its general form, a type-logical grammar consists of following components:

\begin{enumerate}
\item  a \emph{logic}, which fulfils the role of ``Universal
  Grammar'' mainstream linguistics\footnote{This is rather different
    from Montague's use of the term ``Universal
    Grammar'' \citep{ug}. In Montague's sense, the different components
    of a type-logical grammar together
    would be an \emph{instantiation} of Universal Grammar.}, %TODO: add remark of difference with Montague's UG
\item  a \emph{homomorphism} from
  this logic to intuitionistic (linear) logic, this mapping is the
  syntax-semantics interface, with intuitionistic linear logic --- also
  called the Lambek-van Benthem calculus, LP \citep{benthem} ---
  fulfilling the role of ``deep structure''.
\item  a \emph{lexicon}, which is a mapping from words of natural language to sets of formulas in our logic,
\item a set of \emph{goal formulas},  which specifies the formulas corresponding to different types of sentences in our grammar.\footnote{Many authors use a single designated goal formula, typically $s$, as is standard in formal language theory. I prefer this slightly more general setup, since it allows us to distinguish between, for example, declarative sentences, imperatives, yes/no questions, \emph{wh} questions, etc., both syntactically and semantically.}
\end{enumerate}

A sentence $w_1, \ldots, w_n$ is grammatical iff the statement
$A_1,\ldots, A_n \vdash C$ is provable in our logic, for some  $A_i
\in \textit{lex}(w_i)$ and for some goal formula $C$. %where ``$\vdash$''
                                %is the derivability relation in the
                                %given logic. 
In other words, we use the lexicon to map words to formulas and then ask the logic whether the resulting sequence of formulas is a theorem. Parsing in a type-logical grammar is quite literally a form of theorem proving, a very pure realisation of the slogan ``parsing as deduction''.

%\subsection{Architecture}

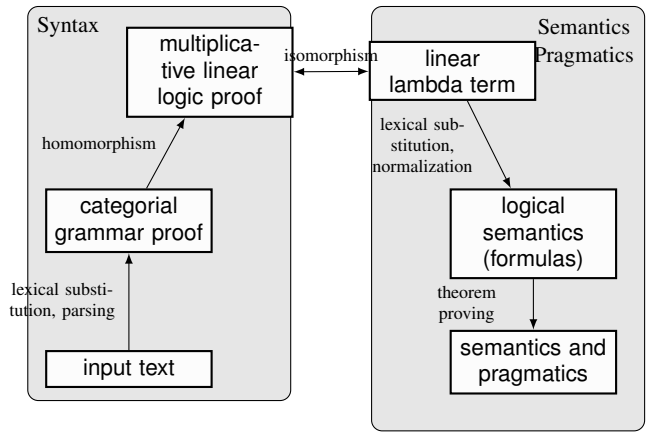
\begin{figure}
\begin{center}
\begin{tikzpicture}[scale=.9,event/.style={rectangle,thick,draw,fill=gray!02,text width=2cm,
		text
                centered,font=\sffamily,anchor=north},label/.style={scale=.8,text
                width=2cm, text
                centered},smalllabel/.style={scale=.55,text
                width=1.5cm,text centered}]
\draw [rounded corners,fill=gray!20] (-5.0em,-9.5em) rectangle (8.0em,10.0em);
\node (syn) at (-3.0em,9.0em) {Syntax};
\draw [rounded corners,fill=gray!20] (12.0em,-11.0em) rectangle (25.5em,10.0em);
\node (sem) at (22.5em,9.0em) {Semantics};
\node (prag) at (22.5em,7.5em) {Pragmatics};
%\draw[line width=0.3cm,color=green!30,cap=round,join=round] (0.5em,-5.5em) -- (17.3em,3em);
%\node [rounded corners,fill=blue!30,draw=black] (lexicon) at
%(10.0em,-0.5em) {\ lexicon\ };
\node [event] (text) at (0em,-7em) {input text};
\node [event] (cg) at (0em,1em) {categorial grammar proof};
\node [event] (mll) at (4em,9em) {mul\-ti\-pli\-ca\-tive linear logic proof};
\node [event] (ll) at (16em,8.3em) {linear lambda term};
\node [event] (sem) at (20em,1em) {logical semantics (formulas)};
\node [event] (prag) at (20em,-6em) {semantics and pragmatics};
\path[>=latex,->] (cg) edge node[left] [label] {{\small homomorphism\qquad
    \qquad \ \ \ \ \ \ \ \qquad \qquad }\ } (mll);
\path[>=latex,<->] (mll) edge node[above] [label] {\small isomorphism} (ll);
\path[>=latex,->] (ll) edge node[left] [label] {\small lexical substitution, normalization} (sem);
\path[>=latex,->] (text) edge node[left] [label] {\small lexical
  substitution, parsing} (cg);
\path[>=latex,->] (sem) edge node[left] [label] {\small theorem
  proving} (prag);
%\node [rounded corners,fill=blue!30,draw=black,text width=2cm,text centered] (ana) at
%(10.0em,-1.2em) {anaphora resolution};
%\path[>=latex,->] (cg) edge node[above] {} %[smalllabel] {c-command}
% (ana);
%\path[>=latex,->] (sem) edge node[above] {} %[smalllabel] {accessibility constraints}
% (ana);
%\path[>=latex,->] (prag) edge node[right] {} %[smalllabel] {
                                %right-frontier constraint}
%(ana);
\end{tikzpicture}
\end{center}
\caption{The standard architecture of type-logical grammars}
\label{fig:arch}
\end{figure}

One of the attractive aspects of type-logical grammars is their simple
and transparent syntax-semantics interface. Though there is a variety
of logics used for
the syntax of type-logical grammars (I will discuss the Lambek
calculus in Section~\ref{sec:lambek} and two generalisations of it in Sections~\ref{sec:mmcg} and~\ref{sec:mill}), there is a large consensus over the syntax-semantics
interface. Figure~\ref{fig:arch} gives a picture of the standard
architecture of type-logical grammars.

%I will present the syntax-semantics interface 

%map to LP, linear lambda term (Curry-Howard)

%Notes about the basic architecture

%\item 

The ``bridge'' between syntax and semantics in the figure is the
Curry-Howard isomorphism between linear lambda terms and proofs in
multiplicative intuitionistic linear logic.
%\item 

Theorem proving occurs in two places of the picture: first
  when parsing a sentence in a given type-logical grammar and also at
  the end when we use the resulting semantics for inferences. I will
  have little to say about this second type of theorem proving
  \citep[provide some investigations into this question, in a way which
  seems compatible with the syntax-semantics interface pursued here,
  though developing a full integrated system combining these systems
  with the current work would be an interesting
  research project]{stergios2015coq,koji2015coq}; theorem proving for parsing will be
  discussed in Section~\ref{sec:parse}.
%\item 

The lexicon plays the role of translating words to syntactic
  formulas but also specifies the semantic term which is used to
  compute the semantics later. The lexicon of a categorial grammar is ``semantically
  informed''. The desired semantics of a sentence allows us to
  reverse-engineer the formula and lexical lambda-term which produce
  it. 
%\item Anaphora resolution requires syntactic information (c-command),
%  semantic information (accessibility conditions) and pragmatic
%  information (right-frontier constraint but also world knowledge).
%\item

Many current semantic theories do not provide a semantic formula
  directly, but first provide a proto-semantics on which further
  computations are performed to produce the final semantics
  (eg.\ for anaphora resolution, presuppositions projection etc.). In
  the current context this means at least some inference is necessary
  to determine semantic and pragmatic wellformedness.
%\end{itemize}

%\subsection{The standard Montagovian picture}

%\subsection{The Montagovian Generative Lexicon}

\subsection{The Lambek calculus}
\label{sec:lambek}

To make things more concrete, I will start by presenting the Lambek
calculus \citep{lambek}. Lambek introduced his calculus as a way to
``obtain an effective rule (or algorithm) for distinguishing sentences
from nonsentences'', which would be applicable both to formal and to (at least fragments of) natural languages \citep[p.\ 154]{lambek}. The simplest formulas used in the Lambek calculus are atomic formulas, which normally include $s$ for sentence, $n$ for common noun, $np$ for noun phrase. We then inductively define the set of formulas of the Lambek calculus by saying that, they include the atomic formulas, and that, if  $A$ and $B$ are formulas (atomic or not), then $A/B$, $A\bullet B$ and $B\bs A$ are also formulas. 

The intended meaning of a formula $A/B$ --- called $A$ over $B$ --- is that
it is looking for an expression of syntactic type $B$ to its right to
produce an expression of syntactic type $A$. An example would be a
word like ``the'' which is assigned the formula $np/n$ in the lexicon,
indicating that it is looking for a common noun (like ``student'') to
its right to form a noun phrase, meaning ``the student'' would be
assigned syntactic type $np$. Similarly, the intended meaning of a
formula $B\bs A$ --- called $B$ under $A$ --- is that it is looking
for an expression of syntactic type $B$ to its left to produce an
expression of type $A$.  This means an intransitive verb like
``slept'', when assigned the formula $np\bs s$ in the lexicon,
combines with a noun phrase to its left to form a sentence $s$. We
therefore predict that ``the student slept'' is a sentence, given the
earlier assignment of $np$ to ``the student''. Finally, a formula $A
\bullet B$ denotes the concatenation of an expression of type $A$ to
an expression of type $B$.

\editout{
Table~\ref{sequentl} lists the sequent calculus rules for the Lambek
calculus. $\Gamma$, $\Delta$, etc.\ denote non-empty sequences of formulas. %The rule $\textit{R} \bullet$ has the condition that neither $\Gamma$ nor $\Delta$ is the empty sequence; the rules $\textit{R} /$ and $\textit{R} \bs$ have the condition that $\Gamma$ is not the empty sequence.

\begin{table}
\begin{center}
\begin{tabular}{cc}
\multicolumn{2}{c}{\bf Identity} \\[3mm]
\infer[\bo \textit{Ax}\bc]{A \ra A}{} &
\infer[\bo \textit{Cut}\bc]{\Gamma ,\Delta,\Gamma^{\prime} \ra C}
                 {\Delta \ra A & \Gamma, A, \Gamma^{\prime} \ra C}\\[3mm]
\end{tabular}

\begin{tabular}{cc}
\multicolumn{2}{c}{\bf Logical Rules} \\[3mm]
\infer[\bo \textit{L}\bullet \bc]{\Gamma, A\bullet B, \Delta \ra C}
                 {\Gamma, A, B, \Delta \ra C} &
\infer[\bo \textit{R}\bullet \bc]{\Gamma,\Delta \ra A\bullet B}
                 {\Gamma \ra A & \Delta \ra B} \\[3mm]
%\end{tabular}
%\begin{tabular}{cc}
\infer[\bo \textit{L}/ \bc]{\Gamma ,A/ B,\Delta,\Gammap \ra C}
                 {\Delta \ra B & \Gamma, A, \Gammap \ra C} &
\infer[\bo \textit{R}/ \bc]{\Gamma \ra A/B}
                 {\Gamma , B \ra A} \\[3mm]
\infer[\bo \textit{L}\bs \bc]{\Gamma, \Delta, B\bs A, \Gammap \ra C}
                   {\Delta \ra B & \Gamma, A, \Gammap \ra C} & 
\infer[\bo \textit{R}\bs \bc]{\Gamma \ra B\bs A}
                   {B, \Gamma \ra A} \\
\end{tabular}
\end{center}
\caption{The sequent calculus \textbf{L}}
\label{sequentl}
\end{table}
}

\begin{table}
\begin{center}
%\begin{tabular}{cc}
%\multicolumn{2}{c}{\bf Identity} \\[3mm]
%\infer[\bo \textit{Ax}\bc]{A \ra A}{} &
%\end{tabular}

\begin{tabular}{cc}
%\multicolumn{2}{c}{\bf Logical Rules} \\[3mm]
\infer[\bo \bullet\textit{E} \bc]{\Gamma, \Delta,\Gammap \ra C}
                 {\Delta\ra A\bullet B & \Gamma, A, B, \Gammap \ra C} &
\infer[\bo \bullet\textit{I} \bc]{\Gamma,\Delta \ra A\bullet B}
                 {\Gamma \ra A & \Delta \ra B} \\[3mm]
%\end{tabular}
%\begin{tabular}{cc}
\infer[\bo /\textit{E} \bc]{\Gamma ,\Delta \ra A}
                 {\Gamma \ra A/B & \Delta \ra B} &
\infer[\bo /\textit{I} \bc]{\Gamma \ra A/B}
                 {\Gamma , B \ra A} \\[3mm]
\infer[\bo \bs\textit{E} \bc]{\Gamma, \Delta \ra A}
                   {\Gamma \ra B & \Delta  \ra B\bs A} & 
\infer[\bo \bs\textit{I} \bc]{\Gamma \ra B\bs A}
                   {B, \Gamma \ra A} \\
\end{tabular}
\end{center}
\caption{Natural deduction for \textbf{L}}
\label{ndl}
\end{table}

Basic statements of the Lambek calculus are of the form $A_1,\ldots,A_n \vdash C$ (with $n
\geq 1$), indicating a claim that the sequence of formulas
$A_1,\ldots, A_n$ is of type $C$; the sequent comma `,' is implicitly
associative and non-commutative.  %TODO: add note derivability
Table~\ref{ndl} shows the natural deduction rules for the Lambek
calculus. $\Gamma$, $\Delta$, etc.\ denote non-empty sequences of
formulas. 

A simple Lambek calculus lexicon is shown in Table~\ref{lambeklex}. I
have adopted the standard convention in type-logical grammars of not
using set notation for the lexicon, but instead listing multiple
lexical entries for a word separately. This corresponds to treating
$\textit{lex}$ as a non-deterministic function rather than as a
set-valued function.

\begin{table}
\begin{align*}
\textit{lex}(\textit{Alyssa}) & = np & \textit{lex}(\textit{ran}) & = np\bs s\\
\textit{lex}(\textit{Emory}) & = np & \textit{lex}(\textit{slept}) & = np\bs s  \\
\textit{lex}(\textit{logic}) & = np & \textit{lex}(\textit{loves}) & = (np\bs s)/np \\
\textit{lex}(\textit{the}) & = np/n & \textit{lex}(\textit{aced}) & = (np\bs s)/np \\
\textit{lex}(\textit{difficult}) & = n/n & \textit{lex}(\textit{passionately}) & = (np\bs s)\bs (np\bs s)  \\
\textit{lex}(\textit{erratic}) & = n/n & \textit{lex}(\textit{during}) & = ((np\bs s)\bs (np\bs s))/np\\
\textit{lex}(\textit{student}) & = n & \textit{lex}(\textit{everyone}) & = s/(np\bs s) \\
\textit{lex}(\textit{exam}) & = n & \textit{lex}(\textit{someone}) & = (s/np)\bs s\\
\textit{lex}(\textit{who}) & = (n\bs n)/(np\bs s)  & \textit{lex}(\textit{every}) & = (s/(np\bs s))/n \\
\textit{lex}(\textit{whom}) & = (n\bs n)/(s / np) & \textit{lex}(\textit{some}) & = ((s/np)\bs s)/n  \\
\end{align*}
\caption{Lambek calculus lexicon}
\label{lambeklex}
\end{table}

Proper names, such as ``Alyssa'' and ``Emory'' are assigned the
category $np$. Common nouns, such as ``student'' and ``exam'' are
assigned the category $n$. Adjectives, such as ``difficult'' or
``erratic'' are not assigned a basic syntactic category but rather the
category $n/n$, indicating they are looking for a common noun to their
right to form a new common noun, so we predict that both ``difficult
exam'' and ``exam'' can be assigned category $n$. For more complex
entries, ``someone'' is looking to its right for a verb phrase to
produce a sentence, where $np\bs s$ is the Lambek calculus equivalent
of verb phrase, whereas ``whom'' is first looking to its right for a
sentence which is itself missing a noun phrase to its right and then
to its left for a noun.

Given the lexicon of Table~\ref{lambeklex}, we can already derive some
fairly complex sentences, such as the following, and, as we will see
in the next section, obtain the correct semantics.

\ex.\label{ex:quant} Every student aced some exam.

\ex. The student who slept during the exam loves Alyssa.

One of the two derivations of Sentence~\ref{ex:quant} is shown in
Figure~\ref{fig:aced}. To improve readability, the figure uses a ``sugared'' notation: instead of writing the lexical
hypothesis corresponding to ``exam'' as $n \vdash n$, we have written
it as $\textit{exam} \vdash n$. The withdrawn $np$'s corresponding to
the object and the subject are given a labels $p_0$ and $q_0$
respectively; the introduction rules are coindexed with the withdrawn
hypotheses, even though this information can be inferred from the rule
instantiation. 

We can always uniquely reconstruct the antecedent from
the labels. For example, the sugared statement ``$p_0\ \textrm{aced}\
q_0 \vdash s$'' in the proof corresponds
to  $np, (np\bs s)/np, np \vdash s$.

\begin{figure}
\scalebox{.8}{
\infer[/ E]{\textrm{every}\ \textrm{student}\ \textrm{aced}\ \textrm{some}\ \textrm{exam} \vdash \textit{s}}{
      \infer[/ E]{\textrm{every}\ \textrm{student} \vdash \textit{s} / (\textit{np} \backslash \textit{s})}{
            \textrm{every}\vdash (\textit{s} / (\textit{np} \backslash \textit{s})) / \textit{n}
            &
            \textrm{student}\vdash \textit{n}
      }
      &
      \infer[\backslash I_{2}]{\textrm{aced}\ \textrm{some}\ \textrm{exam} \vdash \textit{np} \backslash \textit{s}}{
            \infer[\backslash E]{p_{0}\ \textrm{aced}\ \textrm{some}\ \textrm{exam} \vdash \textit{s}}{
                  \infer[/ I_{1}]{p_{0}\ \textrm{aced} \vdash \textit{s} / \textit{np}}{
                        \infer[\backslash E]{p_{0}\ \textrm{aced}\ q_{0} \vdash \textit{s}}{
                              [p_{0}\vdash \textit{np}]^{2}
                              &
                              \infer[/ E]{\textrm{aced}\ q_{0} \vdash \textit{np} \backslash \textit{s}}{
                                    \textrm{aced}\vdash (\textit{np} \backslash \textit{s}) / \textit{np}
                                    &
                                    [q_{0}\vdash \textit{np}]^{1}
                              }
                        }
                  }
                  &
                  \infer[/ E]{\textrm{some}\ \textrm{exam} \vdash (\textit{s} / \textit{np}) \backslash \textit{s}}{
                        \textrm{some}\vdash ((\textit{s} / \textit{np}) \backslash \textit{s}) / \textit{n}
                        &
                        \textrm{exam}\vdash \textit{n}
                  }
            }
      }
}
}
\caption{``Every student aced some exam'' with the subject wide scope
  reading.}
\label{fig:aced}
\end{figure}

Although it is easy to verify that the proof of Figure~\ref{fig:aced}
has correctly applied the rules of the Lambek calculus, finding such a
proof from scratch may look a bit complicated (the key steps at the
beginning of the proof involve introducing two $np$ hypotheses and
then deriving $s/np$ to allow the object quantifier to take narrow scope).
We will defer the question ``given a statement $\Gamma \vdash C$, how
do we decide whether or not it is derivable?'' to
Section~\ref{sec:parse} but will first discuss how this proof
corresponds to the following logical formula.

\[
\forall x. [\mathit{student}(x) \Rightarrow \exists
y. [\mathit{exam}(y) \wedge \mathit{ace}(x,y) ] ]
\]

\subsection{The syntax-semantics interface}
\label{sec:synsem}

For the Lambek calculus, specifying the homomorphism to multiplicative
intuitionistic linear logic is easy: we replace the two implications
`$\bs$' and `$/$' by the linear implication `$\multimap$' and
the product `$\bullet$' by the tensor `$\otimes$'. In a statement
$\Gamma\vdash C$, $\Gamma$ is now a multiset of formulas instead of a
sequence. In other words, the sequent comma `,' is now associative,
commutative instead of associative, non-commutative. For the proof of Figure~\ref{fig:aced} of the previous
section, this mapping gives the proof shown in Figure~\ref{fig:acedlp}.

\begin{figure}
\scalebox{.9}{
\infer[\multimap E]{ \textit{s}}{
      \infer[\multimap E]{ (\textit{np}
        \multimap \textit{s})\multimap \textit{s}}{
            \textit{n} \multimap ((\textit{np} \multimap
            \textit{s})\multimap \textit{s}) 
            &
            \textit{n}
      }
      &
      \infer[\multimap I_{2}]{ \textit{np} \multimap \textit{s}}{
            \infer[\multimap E]{ \textit{s}}{
                  \infer[\multimap I_{1}]{
                    \textit{np}\multimap \textit{s} }{
                        \infer[\multimap E]{ \textit{s}}{
                              [ \textit{np}]^{2}
                              &
                              \infer[\multimap E]{ \textit{np} \multimap \textit{s}}{
                                    \textit{np}\multimap (\textit{np} \multimap \textit{s}) 
                                    &
                                    [\textit{np}]^{1}
                              }
                        }
                  }
                  &
                  \infer[\multimap E]{
                    (\textit{np}\multimap \textit{s}) \multimap \textit{s}}{
                        \textit{n}\multimap ((\textit{np}\multimap \textit{s}) \multimap \textit{s}) 
                        &
                        \textit{n}
                  }
            }
      }
}
}
\caption{Deep structure of the derivation of Figure~\ref{fig:aced}.}
\label{fig:acedlp}
\end{figure}

We have kept the order of the premisses of the rules as they were in
Figure~\ref{fig:aced} to allow for an easier comparison. This deep
structure still uses the same atomic formulas as the Lambek calculus,
it just forgets about the order of the formulas and therefore can no
longer distinguish between the leftward looking implication
`$\backslash$' and the rightward looking implication `$/$'.
% ADD: "some" and 

To obtain a semantics in the tradition of \citeasnoun{montague}, we
use the following mapping from syntactic types to semantic types,
using Montague's atomic types $e$ (for \emph{entity}) and $t$ (for
\emph{truth value}).

\begin{align*}
np^* & = e\\
n^* & = e\rightarrow t\\
s^* & = t\\
(A \multimap B)^* & = A^* \rightarrow B^*
\end{align*}
 
Applying this mapping to the deep structure proof of
Figure~\ref{fig:acedlp} produces the intuitionistic proof and the
corresponding (linear) lambda term as shown in Figure~\ref{fig:acedch}

\newcommand{\tevery}{z_0}
\newcommand{\tstudent}{z_1}
\newcommand{\taced}{z_2}
\newcommand{\tsome}{z_3}
\newcommand{\texam}{z_4}

\begin{figure}
\[
\infer[\rightarrow E]{((\tevery\,\tstudent) \,(\lambda x. ((\tsome\,\texam)\,\lambda y. ((\taced\,y)\,x))))^{\textit{t}}}{
      \infer[\rightarrow E]{ (\tevery\,\tstudent)^{(\textit{e} \rightarrow \textit{t})\rightarrow \textit{t}}}{
             \tevery^{(\textit{e} \rightarrow \textit{t}) \rightarrow (\textit{e} \rightarrow \textit{t})\rightarrow \textit{t} }
            &
            \tstudent^{\textit{e} \rightarrow \textit{t}}
      }
      &
      \infer[\rightarrow I_{2}]{\lambda x. ((\tsome\,\texam)\,\lambda y. ((\taced\,y)\,x))^{\textit{e} \rightarrow \textit{t}}}{
            \infer[\rightarrow E]{((\tsome\,\texam)\,\lambda y. ((\taced\,y)\,x))^{\textit{t}}}{
                  \infer[\rightarrow I_{1}]{
                    \lambda y. ((\taced\,y)\,x)^{\textit{e}\rightarrow \textit{t} }}{
                        \infer[\rightarrow E]{ ((\taced\,y)\,x)^{\textit{t}}}{
                              [ x^{\textit{e}}]^{2}
                              &
                              \infer[\rightarrow E]{ (\taced\,y)^{\textit{e} \rightarrow \textit{t}}}{
                                    \taced^{\textit{e}\rightarrow (\textit{e} \rightarrow \textit{t})} 
                                    &
                                    [y^{\textit{e}}]^{1}
                              }
                        }
                  }
                  &
                  \infer[\rightarrow E]{
                    (\tsome\,\texam)^{ (\textit{e}\rightarrow \textit{t}) \rightarrow \textit{t}}}{
                        \tsome^{(\textit{e}\rightarrow \textit{t})\rightarrow (\textit{e}\rightarrow \textit{t}) \rightarrow \textit{t} }
                        &
                        \texam^{\textit{e}\rightarrow \textit{t}}
                  }
            }
      }
}
\]
\caption{Intuitionistic proof and lambda term corresponding to the
  deep structure of Figure~\ref{fig:acedlp}.}
\label{fig:acedch}
\end{figure}

The computed term corresponds to the derivational semantics of the
proof. To obtain the complete meaning, we need to substitute, for each of $\tevery, \ldots, \texam$, the
meaning assigned in the lexicon. 

For example, ``every'' has syntactic type $(s/(np\bs s))/n$ and
its semantic type is $(e\rightarrow t)\rightarrow (e\rightarrow t)\rightarrow
t$. The corresponding lexical lambda term of this type is $\lambda
P^{e\rightarrow t}.\lambda Q^{e\rightarrow t}. (\forall (\lambda x^e. ((\Rightarrow (P\, x)) (Q\, x))))$, with
`$\forall$' a constant of type $(e\rightarrow t)\rightarrow t$ and
`$\Rightarrow$' a constant of type $t\rightarrow (t\rightarrow t)$. In
the more familiar Montague formulation, this lexical term corresponds
to $\lambda
P^{e\rightarrow t}.\lambda Q^{e\rightarrow t}. \forall x. [ (P\, x)
\Rightarrow (Q\,x)]$, where we can see the formula in higher-order
logic we are constructing more clearly. Although the derivational
semantics is a linear lambda term, the lexical term assigned to
``every'' is not, since the variable $x$ has two bound occurrences.

The formula assigned to ``some'' has the same semantic type but a
different term $\lambda
P^{e\rightarrow t}.\lambda Q^{e\rightarrow t}. (\exists (\lambda x^e. ((\wedge (P\, x)) (Q\, x))))$.

The other words are simple, ``exam'' is assigned
$\mathit{exam}^{e\rightarrow t}$, ``student'' is assigned
$\mathit{student}^{e\rightarrow t}$, and ``aced'' is assigned
$\mathit{ace}^{e\rightarrow (e\rightarrow t)}$. 

So to compute the meaning, we start with the 
derivational semantics, repeated below.

\[
((\tevery\,\tstudent) \,(\lambda x. ((\tsome\,\texam)\,\lambda y. ((\taced\,y)\,x))))
\]

Then we substitute the lexical meanings, for $\tevery,\ldots,\texam$.

\begin{align*}
\tevery & :=  \lambda
P^{e\rightarrow t}.\lambda Q^{e\rightarrow t}. (\forall (\lambda x^e. ((\Rightarrow (P\, x)) (Q\, x))))\\
\tstudent &:= \mathit{student}^{e\rightarrow t}\\
\taced & := \mathit{ace}^{e\rightarrow (e\rightarrow t)}\\
\tsome & := \lambda
P^{e\rightarrow t}.\lambda Q^{e\rightarrow t}. (\exists (\lambda x^e. ((\wedge (P\, x)) (Q\, x))))\\
\texam & := \mathit{exam}^{e\rightarrow t}\\
\end{align*}

This produces the following lambda term.

\begin{align*}
((\lambda
P^{e\rightarrow t}.\lambda Q^{e\rightarrow t}. & (\forall (\lambda
                                                 x^e. ((\Rightarrow
                                                 (P\, x)) (Q\,
                                                 x))))\,\mathit{student}^{e\rightarrow
                                                 t}) \\
\,(\lambda x. ((\lambda
P^{e\rightarrow t}.\lambda Q^{e\rightarrow t}. & (\exists (\lambda
                                                 x^e. ((\wedge (P\,
                                                 x)) (Q\,
                                                 x))))\,\mathit{exam}^{e\rightarrow
                                                 t})\\
&\lambda y. ((\mathit{ace}^{e\rightarrow (e\rightarrow t)}\,y)\,x))))
\end{align*}

Finally, when we normalise this lambda term, we obtain the following
semantics for this sentence.

\[
(\forall (\lambda
                                                 x^e. ((\Rightarrow
                                                 (\mathit{student}^{e\rightarrow
                                                 t})\, x)) (\exists (\lambda
                                                 y^e. ((\wedge (\mathit{exam}^{e\rightarrow
                                                 t}\,
                                                 y)) (((\mathit{ace}^{e\rightarrow (e\rightarrow t)}\,y)\,x)))))
\]

This lambda term represents the more readable higher-order logic
formula\footnote{We have used the standard convention in Montague
  grammar of writing $(p\, x)$ as $p(x)$ and $((p\,y)\,x)$ as
  $p(x,y)$, for a predicate symbol $p$.}.

\[
\forall x. [\mathit{student}(x) \Rightarrow \exists
y. [\mathit{exam}(y) \wedge \mathit{ace}(x,y) ] ]
\]

Proofs in the Lambek calculus, and in type-logical grammars are
subsets of the proofs in intuitionistic (linear) logic and these
proofs are compatible with formal semantics in the tradition initiated
by \citeasnoun{montague}.

For the example in this section, we have calculated the semantics
of a simple example in ``slow motion'': many authors assign a lambda
term directly to a proof in their type-logical grammar, leaving the
translation to intuitionistic linear logic implicit.

Given a semantic analysis without a corresponding syntactic proof, we
can try to reverse engineer the syntactic proof. For example, suppose
we want to assign the reflexive ``himself'' the lambda term $\lambda
R^{(e\rightarrow e\rightarrow t)}\lambda x^e. ((R\,x)\, x)$, that is,
a term of type $(e\rightarrow e\rightarrow t)\rightarrow e\rightarrow t$. Then,
using some syntactic reasoning to eliminate implausible candidates like
$(np\multimap n)\multimap n$, the only reasonable deep structure
formula is $(np\multimap np\multimap s)\multimap (np\multimap s)$ and,
reasoning a bit further about which of the implications is left and
right, we quickly end up with the quite reasonable (though far from
perfect) Lambek calculus formula $((np\bs
s)/np)\bs(np\bs s)$.

\subsection{Going further}
\label{sec:further}

Though the Lambek calculus is a beautiful and simple logic and though
it gives a reasonable account of many interesting phenomena on the
syntax-semantics interface, the Lambek calculus has a number of
problems, which I will discuss briefly below. The driving force of
research in type-logical grammars since the eighties has been to find
solutions to these problems and some of these solutions will be the
main theme of the next section.

\paragraph{Formal language theory} The Lambek calculus generates only
context-free languages \citep{pentus97}. There is a rather large consensus that natural
languages are best described by a class of languages at least slightly
larger than the context-free languages. Classical examples of phenomena
better analysed using so-called mildly context-sensitive language
include verb clusters in Dutch and in  Swiss German \citep{weakin,shieber}.

\paragraph{The Syntax-Semantics Interface} Though our example grammar
correctly predicted two readings for Sentence~\ref{ex:quant} above,
our treatment of quantifiers doesn't scale well. For example, if we
want to predict two readings for the following sentence (which is just Sentence~\ref{ex:quant} where ``some'' and ``every'' have exchanged position)

% every student aced some exam
\ex. Some student aced every exam.

\noindent then we need to add an additional lexical entry both for
``some'' and for ``every''; this is easily done, but we end up with two
lexical formulas for both words. However, this would still not be
enough. For example, the following sentence is also grammatical.

\ex.\label{ex:gave} Alyssa gave every student a difficult exam.

\ex.\label{ex:perj} Alyssa believes a student committed perjury.

In Sentence~\ref{ex:gave}, ``every student'' does not occur in a peripheral position, and
though it is possible to add a more complex formula with the correct
behaviour, we would need yet another formula for
Sentence~\ref{ex:perj}. Sentence~\ref{ex:perj} is generally considered
to have two readings: a \emph{de dicto} reading, where Alyssa doesn't
have a specific student in mind (she could conclude this, for example, when two students make
contradictory statements under oath, this reading can be felicitously
followed by ``but she doesn't know which''), and a \emph{de re}
reading where Alyssa believes a specific student perjured. The Lambek
calculus cannot generate this second reading without adding yet
another formula for ``a''.

It seems we are on the wrong track when we need to add a
new lexical entry for each different context in which a quantifier
phrase occurs. Ideally, we would like a \emph{single} formula for
``every'', ``some'' and ``a'' which applied in all these different
cases. 

Another way to see this is that we want to keep the deep structure formula
$n\multimap ((np\multimap s) \multimap s)$ and that we need to replace
the Lambek calculus by another logic such that the correct deep
structures for the desired readings of sentences like~\ref{ex:gave}
and~\ref{ex:perj} are produced.

\paragraph{Lexical Semantics} The grammar above also overgenerates in
several ways. 

\begin{enumerate}
\item ``ace'' implies a (very positive) form of evaluation with respect to
the object. ``aced the exam'' is good, whereas ``aced Emory'', outside
of the context of a tennis match is bad. ``aced logic'' can only mean
something like ``aced the exam for the logic course''.
\item ``during'' and similar temporal adverbs imply its argument is a temporal
interval: ``during the exam'' is good, but ``during the student'' is
bad, and ``during logic'' can only mean something like ``during the
contextually understood logic lecture''
\end{enumerate}

In the literature on semantics,  there has been an influential
movement towards a richer ontology of types (compared to the ``flat''
Montagovian picture presented above) but also towards a richer set of
operations for \emph{combining} terms of specific types, notably allowing type coercions
\citep{Pus95,asher11web}. So an ``exam'' can be ``difficult'' (it subject
matter, or informational content) but also ``take a long time'' (the
event of taking the exam). The theory of semantics outlined in
the previous section needs to be extended if we want to take these
and other observations into account.

\section{Modern type-logical grammars}
\label{sec:mtlg}

We ended the last section with some problems with using the Lambek
calculus as a theory of the syntax-semantics interface. The problems
are of two different kinds.

\begin{enumerate}
\item  The problems of the syntax-semantic
interface, and, in a sense, also those of formal language theory are
problems where the deep structure is correct but our syntactic
calculus cannot produce an analysis mapping to the desired deep
structure. We will present two solutions to these problems in
Sections~\ref{sec:mmcg} and~\ref{sec:mill}.
\item The problems of lexical semantics, on the other hand require a
  more sophisticated type system than Montague's simply typed
  type-logical with basic types $e$ and $t$ and mechanisms like
  coercions which allow us to conditionally ``repair'' certain type
  mismatches in this system. We will discuss a solution to this
  problem in Section~\ref{sec:mgl}.
\end{enumerate}

\subsection{Multimodal grammars}
\label{sec:mmcg}

Multimodal type-logical grammars \citep{Moo11} take the non-associative Lambek
calculus as its base, but allow multiple families of connectives. 

For the basic statements $\Gamma\vdash C$ of the Lambek calculus, we
ask the question whether we can derive formula $C$, the succedent, from a sequence of
formulas $\Gamma$, the antecedent.
In the
multimodal Lambek calculus, the basic objects are labeled binary
trees\footnote{We can also allow unary branches (and, more generally
  n-ary branches) and the corresponding
  logical connectives. The unary connectives $\Diamond$ and $\Box$ are
widely used, but, since they will only play a marginal role in what
follows, I will not present them to keep the current presentation
simple. However, they form an essential part of the analysis of many
phenomena and are consequently available in the implementation.}. The labels come from a separate set of
indices or modes $I$. Multimodal formulas are then of the form $A/_i
B$, $A\bullet_i B$ and $A\bs_i B$, and antecedent terms are of the
form $\p{i}{\Gamma}{\Delta}$, with $i$ an index from $I$ (we have
omitted the outer brackets for the rules, but the operator $\circ_i$
is non-associative). Sequents are still written as $\Gamma
\vdash C$, but $\Gamma$ is now a binary branching, labeled tree with
formulas as its leaves.

Given a set of words $w_1,\ldots,w_n$ and a goal formula $C$, the
question is now: is there a labeled tree $\Gamma$ with formulas
$A_1,\ldots,A_n$ as its yield, such that $\Gamma\vdash C$ is derivable
and $A_i \in \textit{lex}(w_i)$ for all $i$ (the implementation of
Section~\ref{sec:mmpn} will automatically compute such a $\Gamma$).

The rules of multimodal type-logical grammars are shown in
Table~\ref{ndnli}. In the rules, $\Gamma[\Delta]$ denotes an
antecedent tree $\Gamma$ with distinguished subtree $\Delta$ --- the
subtree notation is a non-associative version of the Lambek calculus
antecedent $\Gamma,\Delta,\Gamma'$, where $\Delta$ is a subsequence
instead of a subtree as it is in $\Gamma[\Delta]$.

\editout{
\begin{table}
\begin{center}
\begin{tabular}{cc}
\multicolumn{2}{c}{\bf Identity} \\[3mm]
\infer[\bo \textit{Ax}\bc]{A \ra A}{} &
\infer[\bo \textit{Cut}\bc]{\Gamma [\Delta] \ra C}
                 {\Gamma [B] \ra C & \Delta \ra B} \\[3mm]
\end{tabular}

\begin{tabular}{cc}
\multicolumn{2}{c}{\bf Logical Rules} \\[3mm]
\infer[\bo \textit{L}\bullet \bc]{\Gamma [A\bullet_i B] \ra C}
                 {\Gamma [\p{i}{A}{B}] \ra C} &
\infer[\bo \textit{R}\bullet \bc]{\p{\it i}{\Gamma}{\Delta} \ra A\bullet_i B}
                 {\Gamma \ra A & \Delta \ra B} \\[3mm]
\infer[\bo \textit{L}/ \bc]{\Gamma [\p{i}{A/_i B}{\Delta}] \ra C}
                {\Delta \ra B & \Gamma[A] \ra C} &
\infer[\bo \textit{R}/ \bc]{\Gamma \ra A/_i B}
                 {\p{i}{\Gamma}{B} \ra A} \\[3mm]
\infer[\bo \textit{L}\bs \bc]{\Gamma [\p{i}{\Delta}{B\bs_i A}] \ra C}
                   {\Delta \ra B & \Gamma[A] \ra C} & 
\infer[\bo \textit{R}\bs \bc]{\Gamma \ra B\bs_i A}
                   {\p{i}{B}{\Gamma} \ra A} \\[3mm]
\end{tabular}

{\bf Structural Rules} \\[3mm]

\mbox{\infer[\bo\textit{SR}\bc]{\Gamma[\Xi[\Delta_{\pi_1},\ldots,\Delta_{\pi_n}]]
\vdash C}{\Gamma[\Xi'[\Delta_1,\ldots,\Delta_n]] \vdash C}}
\end{center}

\caption{The sequent calculus $\textbf{NL}_{\calr}$}
\label{sequentnli}
\end{table}
}

\begin{table}
\begin{center}
%\begin{tabular}{cc}
%\multicolumn{2}{c}{\bf Identity} \\[3mm]
%\infer[\bo \textit{Ax}\bc]{A \ra A}{} &
%\infer[\bo \textit{Cut}\bc]{\Gamma [\Delta] \ra C}
%                 {\Gamma [B] \ra C & \Delta \ra B} \\[3mm]
%\end{tabular}

\begin{tabular}{cc}
\multicolumn{2}{c}{\bf Logical Rules} \\[3mm]
\infer[\bo \bullet_i\textit{E} \bc]{\Gamma [\Delta] \ra C}
                 {\Delta\ra A\bullet_i B & \Gamma[\p{i}{A}{B}] \ra C} &
\infer[\bo \bullet_i\textit{I} \bc]{\p{\it i}{\Gamma}{\Delta} \ra A\bullet_i B}
                 {\Gamma \ra A & \Delta \ra B} \\[3mm]
\infer[\bo /_i\textit{E} \bc]{\p{i}{\Gamma}{\Delta} \ra A}
                {\Gamma \ra A/_i B & \Delta \ra B} &
\infer[\bo /_i\textit{I} \bc]{\Gamma \ra A/_i B}
                 {\p{i}{\Gamma}{B} \ra A} \\[3mm]
\infer[\bo \bs_i\textit{E} \bc]{\p{i}{\Gamma}{\Delta} \ra A}
                   {\Gamma \ra B & \Delta \ra B\bs_i A} & 
\infer[\bo \bs_i \textit{I} \bc]{\Gamma \ra B\bs_i A}
                   {\p{i}{B}{\Gamma} \ra A} \\[3mm]
\end{tabular}

{\bf Structural Rules} \\[3mm]

\mbox{\infer[\bo\textit{SR}\bc]{\Gamma[\Xi[\Delta_{\pi_1},\ldots,\Delta_{\pi_n}]]
\vdash C}{\Gamma[\Xi'[\Delta_1,\ldots,\Delta_n]] \vdash C}}
\end{center}

\caption{Natural deduction for $\textbf{NL}_{\calr}$}
\label{ndnli}
\end{table}

Each logical connective with mode $i$ uses a structural connective
$\circ_i$ in its rule.  For the $/ E$, $\bullet I$ and $\bs E$ rules,
reading from premisses to conclusions, we \emph{build} structure. For
the $/I$, $\bullet E$ and $\bs I$ rules we \emph{remove} a
structural connective with the same mode as the logical
connective. The natural deduction rules use explicit antecedents,
although, for convenience, we will again use coindexation
between the introduction rules for the
implications `$/$' and `$\bs$' and its withdrawn
premiss (and similarly for the $\bullet E$ rule and its two
premisses).
% However, it will sometimes be convenient use explicit
%coindexation, since this allows us to see easily which premisses in a
%proof of a multimodal grammar correspond to the words in the lexicon.

The main advantage of adding modes to the logic is that modes allow us
to control the application of structural rules lexically. This gives
us fine-grained control over the structural rules in our logic.

For example, the base logic is non-associative. Without structural
rules, the sequent $a/b, b/c \vdash a/c$, which is derivable in the
Lambek calculus is \emph{not} derivable in its multimodal incarnation $a/_a b, b/_a c \vdash a/_a c$.
The proof attempt below, with the failed rule application marked by
the `\Lightning' symbol, shows us that the elimination rules and the
introduction rule for this sequent do not match up correctly.

\[
\infer[\bo/ I\bc]{\p{a}{a/_ab}{b/_ac}\vdash a/_a c
}{\infer[\text{\Lightning}]{\p{a}{(\p{a}{a/_ab}{b/_ac})}{c} \vdash
    a}{\infer[\bo/ E\bc]{\p{a}{a/_a b}{(\p{a}{b/_a c}{c})}\vdash
      a}{a/_a b\vdash a/_a b & \infer[\bo/ E\bc]{\p{a}{b/_a c}{c} \vdash b}{b/_a
        c \vdash b/_a c & c\vdash c}}}}
\]

This is where the structural rules, shown at the bottom of
Table~\ref{ndnli} come in. The general form, read from top to bottom, states that we take a
structure $\Gamma$ containing a distinguished subtree $\Xi$ which
itself has $n$ subtrees $\Delta_1,\ldots,\Delta_n$, and we replace
this subtree $\Xi$ with a subtree $\Xi'$ which has the same number of
subtrees, though not necessarily in the same order ($\pi$ is a
permutation on the leaves). In brief, we replace a subtree $\Xi$ by another
subtree $\Xi'$ and possibly rearrange the leaves (subtrees) of $\Xi$,
without deleting or copying any subtrees. Examples of structural rules
are the following.

\begin{center}
\begin{tabular}{ccc}
\infer[\textit{Ass}]{\Gamma[\p{a}{(\p{a}{\Delta_1}{\Delta_2})}{\Delta_3}]
  \vdash C}{\Gamma[\p{a}{\Delta_1}{(\p{a}{\Delta_2}{\Delta_3})}]\vdash
  C} & &
\infer[\textit{MC}]{\Gamma[\p{1}{(\p{0}{\Delta_1}{\Delta_2})}{\Delta_3}]
  \vdash C}{\Gamma[\p{0}{(\p{1}{\Delta_1}{\Delta_3})}{\Delta_2}]
  \vdash C}
\end{tabular}
\end{center}

The first structural rule is one of the structural rules for
associativity. It is the simplest rule which will make the proof
attempt above valid (with $\Gamma[]$ the empty context, $\Delta_1 = a/_a
b$, $\Delta_2 = b/_a c$ and $\Delta_3 = c$). This structural rule
keeps the order of the $\Delta_i$ the same. 

The rule above on the right is
slightly more complicated. There, the positions of $\Delta_2$ and $\Delta_3$
are swapped as are the relative positions of modes 0 and 1. Rules like
this are called ``mixed commutativity'', they permit controlled access
to permutation. One way to see this rule, seen from top to bottom, is
that is ``moves out'' a $\Delta_3$ constituent which is on the right
branch of mode 1. Rules of this kind are part of the solution to
phenomena like Dutch verb clusters \citep{mo94}.

Many modern type-logical grammars, such as the Displacement calculus
and NL$_{cl}$ can be seen as multimodal grammars \citep{ov14,bs14cont}.

\subsection{First-order linear logic}
\label{sec:mill}

We have seen that multimodal type-logical grammars generalise the
Lambek calculus by offering the possibility of fine-tuned controlled
over the application of structural rules. In this section, I will
introduce a second way of extending the Lambek calculus.

Many parsing
algorithms use pairs of integers to represent the start and end
position of substrings of the input string. For example, we can
represent the sentence

\ex.\label{ex:dictore} Alyssa believes someone committed perjury.

\noindent as follows (this is a slightly simplified version of
Sentence~\ref{ex:perj} from Section~\ref{sec:further}); we have
treated ``committed perjury'' as a single word.

\begin{center}
\begin{tikzpicture}[node distance=5em]
\node (0) {0};
\node (1) [right of=0]{1};
\node (2) [right of=1]{2};
\node (3) [right of=2]{3};
\node (4) [node distance=10em, right of=3]{4};
\path (0) edge node[above] [label] {Alyssa} (1);
\path (1) edge node[above] [label] {believes$_{\rule{0pt}{1ex}}$} (2);
\path (2) edge node[above] [label] {someone$_{\rule{0pt}{1ex}}$} (3);
\path (3) edge node[above] [label] {committed perjury$_{\rule{0pt}{1ex}}$} (4);
\end{tikzpicture}
\end{center}

The basic idea of first-order linear logic as a type-logical grammar
is that we can code strings as pairs (or, more generally,
tuples) of integers representing string positions. So for deciding
the grammaticality of a sequence of words $w_1,\ldots, w_n \vdash C$,
with a goal formula $C$, we now give a parametric translation from $\|
A_i \|^{i-1,i}$ for each lexical entry $w_i$ and $\| C\|^{0,n}$ for
the conclusion formula.

Given these string positions, we can assign the noun phrase ``Alyssa''
the formula $np(0,1)$, that is a noun phrase from position 0 to
position 1. The verb ``believes'', which occurs above between position
1 and 2, can then be assigned the complex formula $\forall x_2. [
s(2,x_2) \multimap \forall x_1. [ np(x_1,1) \multimap s(x_1,x_2)] ]$,
meaning that it first selects a sentence to its right (that is,
starting at its right edge, position 2 and ending anywhere) and then a
noun phrase to its left (that is, starting anywhere and ending at its
left edge, position 1) to produce a sentence from the left position of
the noun phrase argument to the right position of the sentence
argument.

We can systematise this translation, following \citeasnoun{mill1}, and
obtain the following translation from Lambek calculus formulas to
first-order linear logic formulas.

\begin{align*}
\| p \|^{x,y} & = p(x,y) \\
\| A / B \|^{x,y} &= \forall z. \| B \|^{y,z} \multimap \| A \|^{x,z}
  \\
\| B\bs A \|^{y,z} &= \forall x. \| B \|^{x,y} \multimap \| A \|^{x,z}
  \\
\| A \bullet B \|^{x,z} &= \exists y. \| A \|^{x,y} \otimes \| B
  \|^{y,z}  
\end{align*}

Given this translation, the lexical entry for ``believes'' discussed
above is simply the translation of the Lambek calculus formula $(np\bs
s)/s$, with position pair $1,2$, to first-order linear logic. Doing
the same for ``committed perjury'' with formula $np\bs s$ and positions $3,4$
gives $\forall z. [np(z,3) \multimap s(z,4)]$. For ``someone'' we
would simply translate the Lambek calculus formula $s/(np\bs s)$, but
we can do better than that: when we translate ``someone'' as $\forall
y_1. \forall y_2. [ (np(2,3) \multimap s(y_1,y_2)) \multimap
s(y_1,y_2) ]$, we improve upon the Lambek calculus analysis.

As we noted in Section~\ref{sec:further}, the Lambek calculus cannot
generate the ``de re'' reading, where the existential quantifier has
wide scope. Figure~\ref{fig:perj} shows how the simple first-order
linear logic analysis \emph{does} derive this reading.

\begin{figure}
\scalebox{.7}{
\infer[\multimap E]{\textrm{s}(0, 4)}{
      \infer[\multimap I_{1}]{\textrm{np}(2, 3) \multimap \textrm{s}(0, 4)}{
            \infer[\multimap E]{\textrm{s}(0, 4)}{
                  \textrm{np}(0, 1)                   &
                  \infer[\forall E]{\textrm{np}(0, 1) \multimap \textrm{s}(0, 4)}{
                        \infer[\multimap E]{\forall C. [\textrm{np}(C, 1) \multimap \textrm{s}(C, 4)]}{
                              \infer[\multimap E]{\textrm{s}(2, 4)}{
                                    [\textrm{np}(2, 3)]^{1}                                     &
                                    \infer[\forall E]{\textrm{np}(2, 3) \multimap \textrm{s}(2, 4)}{
                                          \forall A. [\textrm{np}(A, 3) \multimap \textrm{s}(A, 4)]                                     }
                              }
                              &
                              \infer[\forall E]{\textrm{s}(2, 4) \multimap \forall C. [\textrm{np}(C, 1) \multimap \textrm{s}(C, 4)]}{
                                    \forall B. [\textrm{s}(2, B) \multimap \forall C. [\textrm{np}(C, 1) \multimap \textrm{s}(C, B)]]                               }
                        }
                  }
            }
      }
      &
      \infer[\forall E]{(\textrm{np}(2, 3) \multimap \textrm{s}(0, 4)) \multimap \textrm{s}(0, 4)}{
          \infer[\forall E]{\forall E. [(\textrm{np}(2, 3) \multimap
            \textrm{s}(0, E)) \multimap \textrm{s}(0, E)]}{
                \forall D. \forall E. [(\textrm{np}(2, 3) \multimap \textrm{s}(D, E)) \multimap \textrm{s}(D, E)]}}}
}
\caption{``De re'' reading for the sentence ``Alyssa believes someone
  committed perjury''.}
\label{fig:perj}
\end{figure}

\editout{
\begin{table}
\begin{center}
\begin{tabular}{cc}
\multicolumn{2}{c}{\bf Identity} \\[3mm]

\infer[\bo Ax\bc]{A \vdash A}{} &

\infer[\bo Cut\bc]{\Gamma,\Delta \vdash C}{\Gamma \vdash A & \Delta,A
\vdash C} \\[2mm]
\end{tabular}

\begin{tabular}{cc}
\multicolumn{2}{c}{\bf Logical Rules} \\[3mm]
\infer[\bo L\otimes\bc]{\Gamma,A\otimes B \vdash C}{\Gamma,A,B \vdash
C} &

\infer[\bo R\otimes\bc]{\Gamma,\Delta \vdash A\otimes B}{\Gamma \vdash
A & \Delta \vdash B} \\[2mm]

\infer[\bo L\lolli\bc]{\Gamma,\Delta,A\lolli B\vdash C}{\Delta \vdash
A & \Gamma,B \vdash C} &

\infer[\bo R\lolli\bc]{\Gamma \vdash A\lolli B}{\Gamma,A \vdash B} \\[2mm]
%\end{tabular}

%\begin{tabular}{cc}
\infer[\bo L\exists\bc]{\Gamma, \exists x.A \vdash C}{\Gamma, A
\vdash C} &

\infer[\bo R\exists\bc]{\Gamma \vdash \exists x.A}{\Gamma \vdash A[x:=t]} \\[2mm]

\infer[\bo L\forall\bc]{\Gamma, \forall x.A \vdash C}{\Gamma, A[x:=t] \vdash
C} &

\infer[\bo R\forall\bc]{\Gamma \vdash \forall x.A}{\Gamma \vdash
A} \\

\end{tabular}
\end{center}
\caption{The sequent calculus \textbf{MILL1}}
\label{seqmill1}
\end{table}}

\begin{table}
$$
\begin{array}{ccc}
\infer[\otimes E_i]{C}{A \otimes B & \infer*{C}{[A]^i[B]^i}}
&&
\infer[\otimes I]{A\otimes B}{A & B} \\
\\
\infer[\multimap E]{B}{A & A \multimap B} &&
\infer[\multimap I]{A \multimap B}{\infer*{B}{[A]^i}} \\
\\
\infer[\exists E_i^*]{C}{\exists x .A & \infer*{C}{[A]^i}} &&
\infer[\exists I]{\exists x.A}{A[x:=t]} \\
\\
\infer[\forall E]{A[x:=t]}{\forall x. A} & &
\infer[\forall I^*]{\forall x. A}{A} \\
\end{array}
$$

\vspace{.8ex}
\begin{center}
$\rule{0pt}{1ex}^*$ no free occurrences of $x$ in any of the free hypotheses
\end{center}
%(except the cancelled hypothesis $A$ for the $\exists E$ rule
\caption{Natural deduction rules for MILL1}
\label{tab:millnd}
\end{table}

Besides the Lambek calculus, first-order linear logic has many other
modern type-logical grammars as fragments. Examples include lambda
grammars \citep{oehrle}, the Displacement calculus
\citep{mvf11displacement} and hybrid type-logical grammars \citep{kl12gap}.
We can see first-order linear logic as a sort of ``machine language''
underlying these different formalisms, with each formalism introducing
its own set of abbreviations convenient for the grammar writer. Seeing
first-order linear logic as an underlying language allows us to
compare the analyses proposed for different formalisms and find, in
spite of different starting points, a lot
of convergence. In addition, as discussed in Section~\ref{sec:millpn}, we can
use first-order linear logic as a uniform proof strategy for these formalisms.

\paragraph{Syntax-Semantics Interface} As usual, we obtain the deep
structure of a syntactic derivation by defining a homomorphism from
the syntactic proof to a proof in multiplicative intuitionistic linear
logic. For first-order linear logic, the natural mapping simply
forgets all first-order quantifiers and replaces all atomic predicates
$p(x_1,\ldots,x_n)$ 
by propositions $p$. Since the first-order variables have, so far,
only been used to encode string positions, such a forgetful mapping
makes sense.

 However, other solutions are possible. When we add semantically
 meaningful terms to first-order linear logic, the Curry-Howard
 isomorphism for the first-order quantifiers will give us
 dependent types and this provides a natural connection to the work
 using dependent types for formal semantics \citep{ranta91icg,pp10acg,luo12formal,luo15lambek}.

\subsection{The Montagovian Generative Lexicon}
\label{sec:mgl}

In the previous sections, we have discussed two general solutions to
the problems of the syntax-semantics interface of the Lambek
calculus. Both solutions proposed a more flexible syntactic logic. In
this section, we will discuss a different type of added flexibility,
namely in the syntax-semantics interface itself. 

The basic motivating examples for a more flexible composition have been
amply debated in the literature \citep{Pus95,asher11web}. 
Our solution is essentially the one proposed by \citeasnoun{bmr10tt},
called the Montagovian Generative Lexicon. I will only give a brief
presentation of this framework. More details can be found in
Chapter~\chapmeryretore.

Like many other solutions,
the first step consists of splitting Montague's type $e$ for entities
into several types: physical objects, locations, informational
objects, eventualities, etc. Although there are different opinions with
respect to the correct granularity of types
\citep{Pus95,asher11web,luo12types}, nothing much hinges on this for
the present discussion. 

The second key element is the move to the second-order lambda
calculus, system F \citep{G95}, which allows abstraction over
\emph{types} as well as over terms. In our Lambek calculus, the
determiner ``the'' was assigned the
formula $np/n$ and the type of its lexical semantics was therefore
$(e\rightarrow t) \rightarrow e$, which we implement using the $\iota$
operators of type $(e\rightarrow t) \rightarrow e$, which, roughly
speaking, selects a contextually salient entity from (a characteristic
function of) a set. When we replace the single type $e$ by several
different types, we want to avoid listing several separate
syntactically identical by semantically different entries for ``the''
in the lexicon, and therefore assign it a polymorphic term $\Lambda
\alpha. \iota^{(\alpha\rightarrow t)\rightarrow \alpha}$ of type $\Pi
\alpha. ((\alpha\rightarrow t)\rightarrow \alpha)$, quantifying over
\emph{all} types $\alpha$. Though this
looks problematic, the problem is resolved once we realise that only
certain function words (quantifiers, conjunctions like ``and'') are
assigned polymorphic terms and that we simply use universal
instantiation to obtain the value of the quantifier variable. So if
``student'' is a noun of type human, that is of type $h\rightarrow
t$, then ``the student'' will be of type $h$, instantiating $\alpha$
to $h$. Formally, we use $\beta$ reduction as follows (this is
substitution of types instead of terms, substituting type $h$ for $\alpha$).

\[
((\Lambda
\alpha. \iota^{(\alpha\rightarrow t)\rightarrow \alpha})\{h\}\,
\textit{student}^{h\rightarrow t}) =_{\beta} (\iota\,\textit{student})^h
\]

The final component of the Montagovian Generative Lexicon is a set of
lexically specified, optional transformations. In case of a type
mismatch, an optional transformation can ``repair'' the term.

As an example from \citeasnoun{mr11plurals} and \citeasnoun{mmr13plurals}, one of the classic
puzzles in semantics are plurals and collective and distributive
readings. For example, verbs like ``meet'' have collective readings,
they apply to groups of individuals collectively, so we have the
following contrast, where collectives like committees and plurals like
students can meet, but not singular or distributively quantified noun
phrases. The contrast with verbs like ``sneeze'', which force a
distributive reading is clear.

\ex. The committee met.

\ex. All/the students met

\ex. *A/each/the student met.

\ex. All/the students sneezed. 

\ex. A/each/the student sneezed.

In the Montagovian Generative lexicon, we can models these fact as
follows. First, we assign the plural morphology ``-s'' the semantics
$\Lambda \alpha \lambda
P^{\alpha\rightarrow t} \lambda Q^{\alpha\rightarrow t}. | Q | > 1 \wedge \forall
x^{\alpha}. Q(x) \Rightarrow P(x)$, then ``students'' is assigned the
following term $\lambda Q^{h\rightarrow t}. | Q | > 1 \wedge \forall
x^h. Q(x) \Rightarrow \textit{student}(x)$, that is the sets of
cardinality greater than one such that all its members are
students. Unlike ``student'' which was assigned a term of type
$h\rightarrow t$, roughly a property of humans, the plural ``students'' is assigned a term of type
$(h\rightarrow t)\rightarrow t$, roughly a property of sets of
humans. Consequently, the contrast between ``the student'' and ``the
students'' is that the first is of type $h$ (a human) and the second
of type $h\rightarrow t$ (a set of humans) as indicated below.

\medskip
\begin{tabular}{r|l|r}
phrase & syntactic type & lambda-term \\ \hline 
\textit{the student} & $np$  & $(\iota \textit{student})^h$\\
\textit{the students} & $np$  & $(\iota  (\lambda Q^{h\rightarrow t}. | Q | > 1 \wedge \forall
x^h Q(x) \Rightarrow \textit{student}(x)))^{h\rightarrow t}$
\end{tabular}
\medskip

Therefore, the meaning of ``the students'' is the contextually
determined set of humans, from the sets of more than one human such that all of
them are students.

Then we distinguish the verbs ``meet'' and ``sneeze'' as follows, with
the simpler verb ``sneeze'' simply selecting for a human subject and
the collective verb ``meet'' selecting for a set of humans (of
cardinality greater than one) as its subject.

\medskip
\begin{tabular}{r|l|r}
word & syntactic type & lambda-term \\ \hline 
\textit{met} & $np\backslash s$  & $\lambda P^{h\rightarrow t}. | P |
> 1 \wedge \textit{meet}(P)$\\
\textit{\#} & & $\Lambda \alpha \lambda R^{(\alpha \rightarrow t) 
\rightarrow t} \lambda S^{(\alpha\rightarrow t)\rightarrow t} \forall 
P^{\alpha\rightarrow t}. S(P) 
\Rightarrow R(P) $\\
$\textit{met}^{\#}$ & $np\backslash s$ & $\lambda R^{(h\rightarrow t) 
\rightarrow t} \forall P^{h\rightarrow t}. R(P) 
\Rightarrow |P | > 1 \wedge \textit{meet}(P) $\\ \hline
%\end{tabular}
%
%\begin{tabular}{r|l|r}
\textit{sneezed} & $np\backslash s$ & $\lambda x^h.
\textit{sneeze}(x)$ \\
\textit{*} &  & $\Lambda \alpha  \lambda P^{\alpha\rightarrow t} \lambda Q^{\alpha\rightarrow 
t} \forall x^{\alpha}. Q(x) \Rightarrow P(x)$ \\
$\textit{sneezed}^*$ & $np\backslash s$ & $\lambda P^{h\rightarrow 
t}. \forall x^h. P(x) \Rightarrow 
\textit{sneeze}(x)$ \\
\end{tabular}
\medskip

Given these basic lexical entries, we already correctly predict that ``the
student met'' is ill-formed semantically (there is an unresolvable
type mismatch) but ``the students met'' and ``the student sneezed''
are given the correct semantics. 

The interesting case is ``the students sneezed'' which has as its only
reading that each student sneezed individually. Given that ``the
students'' is of type $h\rightarrow t$ and that ``sneezed'' requires an
argument of type $h$, there is a type mismatch when we apply the two
terms. However, ``sneeze'' has the optional distributivity operator
`*', which when we apply it to the lexical semantics for ``sneeze''
produces the term $\lambda P^{h\rightarrow 
t}. \forall x^h. P(x) \Rightarrow 
\textit{sneeze}(x)$, which combines with ``the students'' to produce
the reading.  

\[
\forall x^h. (\iota  (\lambda Q^{h\rightarrow t}. | Q | > 1 \wedge \forall
y^h Q(y) \Rightarrow \textit{student}(y))\, x) \Rightarrow 
\textit{sneeze}(x)
\]

In other words, all of the members of the contextually determined set
of more than human which are all students, sneeze.

The basic idea for the Montagovian Generative Lexicon is that lexical
entries specify optional transformations which can repair certain
sorts of type mismatches in the syntax-semantics interface. This
adaptability allows the framework to solve many semantic puzzles.

Though a proof-of-concept application of these ideas exists, more robust and scalable applications, as well as efforts incorporate these ideas into wide-coverage semantics, are ongoing research.

\section{Theorem proving}
\label{sec:parse}

When looking at the rules and examples for the different logics, the
reader may have wondered: how do we actually find proofs for
type-logical grammars? This question becomes especially urgent once
our grammars become more complex and the consequences of our lexical
entries, given our logic, become hard to oversee. Though pen and paper
generally suffice to show that a given sentence is derivable for the
desired reading, it is generally much more laborious to show that a
given sentence is underivable or that it has only the desired
readings. This is where automated theorem provers are useful: they
allow more extensive and intensive testing of your grammars, producing
results more quickly and with less errors (though we should be careful about too naively assuming the implementation we are using is correct: when a proof is found it
is generally easy to verify its correctness by hand, but when a proof isn't found because of a programming error this can be hard to detect). 

Though the natural deduction calculi we have seen so far can be used
for automated theorem proving \citep{carpenter1994natural,mr12lcg}, and
though \citeasnoun{lambek} already gave a sequent calculus decision
procedure, both logics have important drawbacks for proof search.

Natural deduction proofs have a 1-1 correspondence between proofs and
readings, though this is somewhat complicated to enforce for a logic
with the $\bullet \textit{E}$ rule (and the related $\Diamond \textit{E}$ rule).
For the sequent calculus, the product rule is just like the other
rules, but sequent calculus suffers from the so-called ``spurious ambiguity''
problem, which means that it generates many more proofs than readings.

Fortunately, there are proof systems which combine the good aspects of
natural deduction and sequent calculus, and which eliminate their respective drawbacks. Proof nets are a graphical
representation of proofs first introduced for linear logic
\citep{Girard}. Proof nets
suffer neither from spurious ambiguity nor from complications for the
product rules.

Proof nets are usually defined as a subset of a larger class, called
\emph{proof structures}. Proof structures are ``candidate proofs'':
part of the search space of a naive proof search procedure which need
not correspond to actual proofs. Proof nets are those proof structures which
correspond to sequent proofs. Perhaps surprisingly, we can distinguish proof nets from other proof
structures by looking only at graph-theoretical properties of these structures.

Proof search for type-logical grammars using proof nets uses the
following general procedure.

\begin{enumerate}
\item For each of the words in the input sentence, find one of the
  formulas assigned to it in the lexicon.
\item Unfold the formulas to produce a partial proof structure.
\item Enumerate all proof structures for the given formulas by
  identifying nodes in the partial proof structure.
\item Check if the resulting proof structure is a proof net according
  to the correctness condition.
\end{enumerate}

In Sections~\ref{sec:mmpn} and~\ref{sec:millpn} we will instantiate this general procedure for multimodal type-logical grammar and for first-order linear logic respectively.

\subsection{Multimodal proof nets}
\label{sec:mmpn}

\begin{table}
\begin{center}
\begin{tikzpicture}[scale=0.75]
% /E
\node (labl) at (3em,7.5em) {$[\ldr E]$};
\node (ab) at (3em,9.8em) {$C$};
\node (a) at (0,14.6em) {$C/_i B$};
\node (aa) at (0.7em,14.2em) {};
\node (b) at (6em,14.75em) {$B$};
\node[tns] (c) at (3em,12.668em) {};
\node (clab) at (3em,12.668em) {$i$};
\draw (c) -- (ab);
\draw (c) -- (aa);
\draw (c) -- (b);
% /I
\node (labl) at (3em,-2.0em) {$[\ldr I]$};
\node (pa) at (0,0) {$C/_i B$};
\node (pat) at (0.7em,0.44em) {};
\node[par] (pc) at (3em,1.732em) {};
\node (pclab) at (3em,1.732em) {\textcolor{white}{$i$}};
\node (pb) at (6em,0.15em) {$B$};
\node (pd) at (3em,4.8em) {$C$};
\draw (pc) -- (pb);
\draw (pc) -- (pd);
\path[>=latex,->]  (pc) edge (pat);
% \E
\node (labl) at (23em,7.5em) {$[\ldl E]$};
\node (ab) at (23em,9.8em) {$C$};
\node (a) at (26em,14.6em) {$A\bs_i C$};
\node (aa) at (25.3em,14.2em) {};
\node (b) at (20em,14.75em) {$A$};
\node[tns] (c) at (23em,12.668em) {};
\node (clab) at (23em,12.668em) {$i$};
\draw (c) -- (ab);
\draw (c) -- (aa);
\draw (c) -- (b);
% \I
\node (labl) at (23em,-2.0em) {$[\ldl I]$};
\node (pa) at (26em,0) {$A\bs_i C$};
\node (pat) at (25.3em,0.44em) {};
\node[par] (pc) at (23em,1.732em) {};
\node (pclab) at (23em,1.732em) {\textcolor{white}{$i$}};
\node (pb) at (20em,0.15em) {$A$};
\node (pd) at (23em,4.8em) {$C$};
\draw (pc) -- (pb);
\draw (pc) -- (pd);
\path[>=latex,->]  (pc) edge (pat);
% *E
\node (labl) at (13em,7.5em) {$[\bullet E]$};
\node (pa) at (10em,9.8em) {$A$};
\node (pdt) at (13em,14.3em) {};
\node[par] (pc) at (13em,11.532em) {};
\node (pclab) at (13em,11.532em) {\textcolor{white}{$i$}};
\node (pb) at (16em,9.8em) {$B$};
\node (pd) at (13em,14.6em) {$A\bullet_i B$};
\draw (pc) -- (pb);
\draw (pc) -- (pa);
\path[>=latex,->]  (pc) edge (pdt);
% *I
\node (labl) at (13em,-2.0em) {$[\bullet I]$};
\node (ab) at (13em,0em) {$A\bullet_i B$};
\node (aba) at (13em,0.3em) {};
\node (a) at (16em,4.8em) {$B$};
\node (b) at (10em,4.8em) {$A$};
\node[tns] (c) at (13em,2.868em) {};
\node (clab) at (13em,2.868em) {$i$};
\draw (c) -- (aba);
\draw (c) -- (a);
\draw (c) -- (b);
\end{tikzpicture}
\end{center}
\caption{Links for multimodal proof nets}
\label{tab:mmlinks}
\end{table}

Table~\ref{tab:mmlinks} presents the links for multimodal proof
nets. The top row list the links corresponding to the elimination
rules of natural deduction, the bottom row those corresponding to the
introduction rules. There are two types of links: tensor links, with
an open center, and par links, with a filled center. Par links have a
single arrow pointing to the main formula of the link (the complex
formula containing the principal connective). The top and
bottom row are up-down symmetric with tensor and par reversed.
The tensor links correspond to the logical rules which build structure
when we read them from top to bottom, the par links to those rules
which remove structure.

The formulas written above the central node of a link are its
premisses, whereas the formulas written below it are its
conclusions. Left-to-right order of the premisses as well as the
conclusions is important.

A \emph{proof structure} is a set of formula occurrences and a set of
links such that:

\begin{enumerate}
\item each formula is at most once the premiss of a link,
\item each formula is at most once the conclusion of a link.
\end{enumerate}

A formula which is not the premiss of any link is a conclusion of the
proof structure. A formula which is not the conclusion of any link is
a hypothesis of the proof structure. We say a proof structure with
hypotheses $\Gamma$ and conclusions $\Delta$ is a proof structure of
$\Gamma \vdash \Delta$ (we are overloading of the `$\vdash$' symbol
here, though this use should always be clear from the context;
note that $\Delta$ can contain multiple formulas).

After the first step of lexical lookup we have a sequent $\Gamma\vdash
C$, and we can enumerate its proof
structures as follows: unfold the formulas in $\Gamma, C$, unfolding them so that the formulas in $\Gamma$ are hypotheses and the formula $C$ is a conclusion of the
resulting structure, until we
reach the atomic subformulas (this is step 2 of the general
procedure), then identify atomic subformulas (step 3 of the general
procedure, we turn to the last step, checking correctness, below). This
identification step can, by the conditions on proof structures only
identify hypotheses with conclusions and must leave all formulas of
$\Gamma$, including atomic
formulas, as hypotheses and $C$ as a
conclusion. 

\begin{figure}
\begin{center}
\begin{tikzpicture}[scale=0.75]
% a/b
%\node (labl) at (3em,7.5em) {$[\ldr E]$};
\node (ab) at (3em,9.8em) {$a$};
\node (a) at (0,14.6em) {$a/_a b$};
\node (aa) at (0.7em,14.2em) {};
\node (b) at (6em,14.75em) {$b$};
\node[tns] (c) at (3em,12.668em) {};
\node (clab) at (3em,12.668em) {$a$};
\draw (c) -- (ab);
\draw (c) -- (aa);
\draw (c) -- (b);
% b/c
\node (bbc) at (13em,9.8em) {$b$};
\node (bc) at (10em,14.6em) {$b/_a c$};
\node (aabc) at (10.7em,14.2em) {};
\node (cbc) at (16em,14.75em) {$c$};
\node[tns] (cc) at (13em,12.668em) {};
\node (clab) at (13em,12.668em) {$a$};
\draw (cc) -- (bbc);
\draw (cc) -- (aabc);
\draw (cc) -- (cbc);
% |- a/c
\node (pa) at (20em,9.8em) {$a/_a c$};
\node (pat) at (20.7em,10.24em) {};
\node[par] (pc) at (23em,11.532em) {};
\node (pclab) at (23em,11.532em) {\textcolor{white}{$a$}};
\node (pb) at (26em,9.8em) {$c$};
\node (pd) at (23em,14.6em) {$a$};
\draw (pc) -- (pb);
\draw (pc) -- (pd);
\path[>=latex,->]  (pc) edge (pat);
\end{tikzpicture}
\end{center}
\caption{Lexical unfolding of $a/_a b, b/_a c \vdash a/_a c$}
\label{fig:unfold} 
\end{figure}
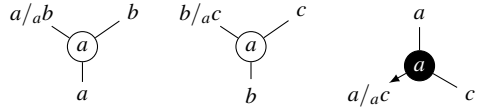

Figure~\ref{fig:unfold} shows the lexical unfolding of the 
sequent $a/_a b, b/_a c \vdash a/_a c$. It is already a proof 
structure, though a proof structure of $a, a/_a b, b, b/_a c, c \vdash 
a, a/_a c, b, c$ (to the reader familiar with the proof nets of linear 
logic: some other presentations of proof nets use more
restricted definitions of proof structures where a ``partial proof
structure'' such as shown in the figure is called a \emph{module}).

\begin{figure}
\begin{center}
\begin{tikzpicture}[scale=0.75]
% a/b
%\node (labl) at (3em,7.5em) {$[\ldr E]$};
\node (ab) at (3em,9.8em) {$a$};
\node (a) at (0,14.6em) {$a/_a b$};
\node (aa) at (0.7em,14.2em) {};
\node (b) at (6em,14.75em) {$b$};
\node[tns] (c) at (3em,12.668em) {};
\node (clab) at (3em,12.668em) {$a$};
\draw (c) -- (ab);
\draw (c) -- (aa);
\draw (c) -- (b);
% b/c
%\node (bbc) at (6em,14.8em) {$b$};
\node (bc) at (3em,19.6em) {$b/_a c$};
\node (aabc) at (3.7em,19.2em) {};
\node (cbc) at (9em,19.75em) {$c$};
\node[tns] (cc) at (6em,17.668em) {};
\node (clab) at (6em,17.668em) {$a$};
\draw (cc) -- (b);
\draw (cc) -- (aabc);
\draw (cc) -- (cbc);
% |- a/c
\node (pa) at (0em,5.0em) {$a/_a c$};
\node (pat) at (0.3em,5.44em) {};
\node[par] (pc) at (3em,6.732em) {};
\node (pclab) at (3em,6.732em) {\textcolor{white}{$a$}};
\node (pb) at (6em,5.0em) {$c$};
%\node (pd) at (3em,9.8em) {$a$};
\draw (pc) -- (pb);
\draw (pc) -- (ab);
\path[>=latex,->]  (pc) edge (pat);
%
% a/b
%\node (labl) at (3em,7.5em) {$[\ldr E]$};
\node (ab) at (18em,9.8em) {$a$};
\node (a) at (15em,14.6em) {$a/_a b$};
\node (aa) at (15.7em,14.2em) {};
\node (b) at (21em,14.75em) {$b$};
\node[tns] (c) at (18em,12.668em) {};
\node (clab) at (18em,12.668em) {$a$};
\draw (c) -- (ab);
\draw (c) -- (aa);
\draw (c) -- (b);
% b/c
%\node (bbc) at (6em,14.8em) {$b$};
\node (bc) at (18em,19.6em) {$b/_a c$};
\node (aabc) at (18.7em,19.2em) {};
\node (cbc) at (24em,19.75em) {$c$};
\node[tns] (cc) at (21em,17.668em) {};
\node (clab) at (21em,17.668em) {$a$};
\draw (cc) -- (b);
\draw (cc) -- (aabc);
\draw (cc) -- (cbc);
% |- a/c
\node (pa) at (15em,5.0em) {$a/_a c$};
\node (pat) at (15.3em,5.44em) {};
\node[par] (pc) at (18em,6.732em) {};
\node (pclab) at (18em,6.732em) {\textcolor{white}{$a$}};
%\node (pb) at (21em,5.0em) {$c$};
%\node (pd) at (3em,9.8em) {$a$};
%\draw (pc) -- (pb);
\draw (pc) -- (ab);
\path[>=latex,->]  (pc) edge (pat);
\draw (cbc) to [out=50,in=330] (pc);
\end{tikzpicture}
\caption{The proof structure of Figure~\ref{fig:unfold} after
  identification of the $a$ and $b$ atoms (left) and after
  identification of all atoms}
\label{fig:ps}
\end{center}
\end{figure}
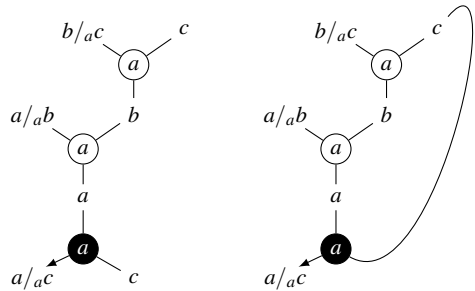

To turn this proof structure into a proof structure of $a/_a b, b/_a c
\vdash a/_a c$, we identify the atomic formulas. In this case, there
is only a single way to do this, since $a$, $b$ and $c$ all occur
once as a hypothesis and once as a conclusion, though in general there
may be many possible matchings. Figure~\ref{fig:ps} shows, on the left, the proof
structure after identifying the $a$ and $b$ formulas. Since left and
right (linear order), up and down (premiss, conclusion) have meaning
in the graph, connecting the $c$ formulas is less obvious: $c$ is a
conclusion of the $/I$ link and must therefore be below it, but a
premiss of the $/E$ link and must therefore be above it. This is hard
to achieve in the figure shown on the left. Though a possible solution
would be to draw the figure on a cylinder, where ``going up'' from the
topmost $c$ we arrive at the bottom one, for ease of type-setting and
reading the figure, I have chosen the representation shown in
Figure~\ref{fig:ps} on the right. The curved line goes up from the $c$
premiss of the $/E$ link and arrives from below at the $/I$ link, as
desired. One way so see this strange curved connection is as a
graphical representation of the coindexation of a premiss with a rule
in the natural deduction rule for the implication.

Figure~\ref{fig:ps} therefore shows, on the right, a proof structure
for $a/_a b, b/_a c
\vdash a/_a c$. However, is it also a \emph{proof net}, that is, does
it correspond to a proof? In a
multimodal logic, the answer depends on the available structural
rules. For example, if no structural rules are applicable to mode $a$
then $a/_a b, b/_a c
\vdash a/_a c$ is underivable, but if mode $a$ is associative, then it
is derivable.

\begin{figure}
\begin{center}
\begin{tikzpicture}[scale=0.75]
% a/b
%\node (labl) at (3em,7.5em) {$[\ldr E]$};
\node (ab) at (18em,9.8em) {$a$};
\node (a) at (15em,14.6em) {$a/_a b$};
\node (aa) at (15.7em,14.2em) {};
\node (b) at (21em,14.75em) {$b$};
\node[tns] (c) at (18em,12.668em) {};
\node (clab) at (18em,12.668em) {$a$};
\draw (c) -- (ab);
\draw (c) -- (aa);
\draw (c) -- (b);
% b/c
%\node (bbc) at (6em,14.8em) {$b$};
\node (bc) at (18em,19.6em) {$b/_a c$};
\node (aabc) at (18.7em,19.2em) {};
\node (cbc) at (24em,19.75em) {$c$};
\node[tns] (cc) at (21em,17.668em) {};
\node (clab) at (21em,17.668em) {$a$};
\draw (cc) -- (b);
\draw (cc) -- (aabc);
\draw (cc) -- (cbc);
% |- a/c
\node (pa) at (15em,5.0em) {$a/_a c$};
\node (pat) at (15.3em,5.44em) {};
\node[par] (pc) at (18em,6.732em) {};
\node (pclab) at (18em,6.732em) {\textcolor{white}{$a$}};
%\node (pb) at (21em,5.0em) {$c$};
%\node (pd) at (3em,9.8em) {$a$};
%\draw (pc) -- (pb);
\draw (pc) -- (ab);
\path[>=latex,->]  (pc) edge (pat);
\draw (cbc) to [out=50,in=330] (pc);
% a/b
%\node (labl) at (3em,7.5em) {$[\ldr E]$};
\node (ab) at (33em,9.8em) {$\apsnodei$};
\node (a) at (30em,14.6em) {$a/_a b$};
\node (aa) at (30.7em,14.2em) {};
\node (b) at (36em,14.75em) {$\apsnodei$};
\node[tns] (c) at (33em,12.668em) {};
\node (clab) at (33em,12.668em) {$a$};
\draw (c) -- (ab);
\draw (c) -- (aa);
\draw (c) -- (b);
% b/c
%\node (bbc) at (6em,14.8em) {$b$};
\node (bc) at (33em,19.6em) {$b/_a c$};
\node (aabc) at (33.7em,19.2em) {};
\node (cbc) at (39em,19.75em) {$\apsnodei$};
\node[tns] (cc) at (36em,17.668em) {};
\node (clab) at (36em,17.668em) {$a$};
\draw (cc) -- (b);
\draw (cc) -- (aabc);
\draw (cc) -- (cbc);
% |- a/c
\node (pa) at (30em,5.0em) {$a/_a c$};
\node (pat) at (30.3em,5.44em) {};
\node[par] (pc) at (33em,6.732em) {};
\node (pclab) at (33em,6.732em) {\textcolor{white}{$a$}};
%\node (pb) at (21em,5.0em) {$c$};
%\node (pd) at (3em,9.8em) {$a$};
%\draw (pc) -- (pb);
\draw (pc) -- (ab);
\path[>=latex,->]  (pc) edge (pat);
\draw (cbc) to [out=50,in=330] (pc);
\end{tikzpicture}
\caption{The proof structure of Figure~\ref{fig:ps} (left) and its
  abstract proof structure (right)}
\label{fig:aps}
\end{center}
\end{figure}
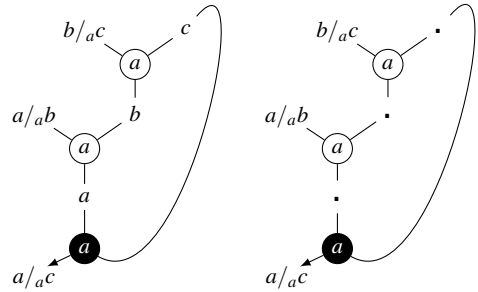

We decide whether a proof structure is a proof net based only on
properties of the graph. As a first step, we erase all formula
information from the internal nodes of the graph; for administrative
reasons, we still need to be able to identify which of the hypotheses
and conclusion of the structure correspond to which formula
occurrence\footnote{We make a slight simplification here. A single
  vertex abstract proof structure can have both a hypothesis and a
  conclusion without these two formulas necessarily being identical,
  e.g.\ for sequents like $(a/b)\bullet b\vdash a$. Such a sequent
  would correspond to the abstract proof structure
  $\overset{(a/b)\bullet b}{\underset{a}{\cdot}}$. So, formally, both the
  hypotheses and the conclusions of an abstract proof structure are
  assigned a formula and when a node is both a hypothesis and a
  conclusion it can be assigned two different formulas. In order not
  to make the notation of abstract proof structure more complex, we
  will stay with the simpler notation. \citeasnoun{mp} present the full details.}. All relevant information for correctness is present in
this graph, which we call an \emph{abstract proof structure}.

We talked about how the curved line in proof structures (and abstract
proof structure) corresponds to the coindexation of discharged
hypotheses with rule names for the implication introduction
rules. However, the introduction rules for multimodal type-logical
grammars actually do more than just discharge a hypothesis, they also
check whether the discharged hypothesis is the immediate left (for
$\bs I$) or right (for $/ I$) daughter of the root node, that is, that
the withdrawn hypothesis $A$ occurs as $A\circ_i \Gamma$ (for $\bs I$
and mode $i$) or $\Gamma\circ_i A$ (for $/I$ and mode $i$). The par
links in the (abstract) proof structure represent a sort of ``promise''
that will produce the required structure. We check whether it is
satisfied by means of contractions on the abstract proof structure.

\begin{table}
\begin{center}
%\begin{tabular}{ccc}
\begin{tikzpicture}[scale=0.75]
\node (labl) at (3em,-2.5em) {$[\ldr I]$};
\node (ab) at (3em,4.8em) {$\apsnodei$};
\node (a) at (0,9.6em) {$\apsnodei$};
\node (b) at (6em,9.6em) {$\apsnodei$};
\node[tns] (c) at (3em,7.668em) {};
\node (clab) at (3em,7.668em) {$i$};
\draw (c) -- (ab);
\draw (c) -- (a);
\draw (c) -- (b);
\node (pa) at (0,0) {$\apsnodei$};
\node[par] (pc) at (3em,1.732em) {};
\node (pclab) at (3em,1.732em) {\textcolor{white}{$i$}};
\draw (pc) -- (ab);
\path[>=latex,->]  (pc) edge (pa);
\draw (b) to [out=50,in=330] (pc);
%\end{tikzpicture}
%&
%\begin{tikzpicture}
\node (labm) at (13em,-2.5em) {$[\lpr E ]$};
\node (tab) at (13em,0.0em) {$\apsnodei$};
\node (ta) at (10em,4.8em) {$\apsnodei$};
\node (tb) at (16em,4.8em) {$\apsnodei$};
\node[tns] (tc) at (13em,2.868em) {};
\node (tclab) at (13em,2.868em) {$i$};
\draw (tc) -- (tab);
\draw (tc) -- (ta);
\draw (tc) -- (tb);
\node (tabx) at (13em,9.6em) {$\apsnodei$};
\node[par] (tcx) at (13em,6.532em) {};
\node (tcxlab) at (13em,6.532em) {\textcolor{white}{$i$}};
\path[>=latex,->]  (tcx) edge (tabx);
\draw (tcx) -- (ta);
\draw (tcx) -- (tb);
%\end{tikzpicture} 
%&
%\begin{tikzpicture}
\node (labr) at (23em,-2.5em) {$[\ldl I ]$};
\node (ab) at (23em,4.8em) {$\apsnodei$};
\node (a) at (26em,9.6em) {$\apsnodei$};
\node (b) at (20em,9.6em) {$\apsnodei$};
\node[tns] (c) at (23em,7.668em) {};
\node (clab) at (23em,7.668em) {$i$};
\draw (c) -- (ab);
\draw (c) -- (a);
\draw (c) -- (b);
\node (pa) at (26em,0) {$\apsnodei$};
\node[par] (pc) at (23em,1.732em) {};
\node (pclab) at (23em,1.732em) {\textcolor{white}{$i$}};
\draw (pc) -- (ab);
\path[>=latex,->]  (pc) edge (pa);
\draw (b) to [out=130,in=210] (pc);
\end{tikzpicture}
%\end{tabular}
\end{center}
\caption{Contractions --- multimodal binary connectives}
\label{fig:lambek_contr}
\end{table}

The multimodal contractions are shown in
Table~\ref{fig:lambek_contr}. All portrayed configurations contract
to a single vertex: we erase the two internal vertices and the paired
links and we identify the two external vertices, keeping all
connections of the external vertices to the rest of the abstract proof
structure as they were: the vertex which is the result of the
contraction will be a conclusion of the same link as the top
external vertex (or a hypothesis of the abstract proof structure in
case it wasn't) and it will be a premiss of the same link as the
bottom external vertex (or a conclusion of the abstract proof
structure in case it wasn't).

The contraction for $/I$ checks if the withdrawn hypothesis is the
right daughter of a tensor link with the same mode information
$i$, and symmetrically for the $\bs I$ contraction. The $\bullet E$
contraction contracts two hypotheses occurring as sister nodes.

All contractions are instantiations of the same pattern: a tensor
link and a par link are connected, respecting left-right and up-down
the two vertices of the par link without the arrow.

To get a better feel for the contractions, we will start with its
simplest instances. When we do pattern matching on the contraction for $/ I$, we see that
it corresponds to the following patterns, depending on our choice for
the tensor link (the par link is always $/ I$).

\begin{align*}
C/_i B &\vdash C/_i B \\
A & \vdash (A\bullet_i B)/_i B \\
A & \vdash C/_i (A\backslash_i C) 
\end{align*}

A proof structure is a proof net iff it contracts to a tree containing
only tensor links using the contractions of Table~\ref{fig:lambek_contr} and any structural rewrites, discussed
below --- \citeasnoun{mp} present full proofs. In other words, we need to contract all par links in the proof
structure according to their contraction, each contraction ensuring
the correct application of the rule after which it is named. The
abstract proof structure on the right of Figure~\ref{fig:aps} does not
contract, since there is no substructure corresponding to the $/I$
contraction: for a valid contraction, a par link is
connected to both ``tentacles'' of a single tensor link, and in the figure the two
tentacles without arrow are connected to different tensor links.
This is correct, since $a/_a b, b/_a c\vdash a/_a c$ is underivable in
a logic without structural rules for $a$.

\begin{figure}
\begin{center}
\begin{tikzpicture}[scale=0.75]
% a/b 
%\node (labl) at (3em,7.5em) {$[\ldr E]$};
\node (ab) at (13em,9.8em) {$v$};
\node (a) at (10em,14.6em) {$x$};
\node (aa) at (10.7em,14.2em) {};
\node (b) at (16em,14.75em) {$\apsnodei$};
\node[tns] (c) at (13em,12.668em) {};
\node (clab) at (13em,12.668em) {$a$};
\draw (c) -- (ab);
\draw (c) -- (aa);
\draw (c) -- (b);
% b/c 
%\node (bbc) at (6em,14.8em) {$b$};
\node (bc) at (13em,19.6em) {$y$};
\node (aabc) at (13.7em,19.2em) {};
\node (cbc) at (19em,19.6em) {$z$};
\node[tns] (cc) at (16em,17.668em) {};
\node (clab) at (16em,17.668em) {$a$};
\draw (cc) -- (b);
\draw (cc) -- (aabc);
\draw (cc) -- (cbc);
%
% a/b 
%\node (labl) at (3em,7.5em) {$[\ldr E]$};
\node (ab) at (34em,9.8em) {$v$};
\node (a) at (31em,14.6em) {$\apsnodei$};
%\node (a) at (31em,14.6em) {$a/_a b$};
%\node (aa) at (31.7em,14.2em) {};
\node (b) at (37em,14.6em) {$z$};
\node[tns] (c) at (34em,12.668em) {};
\node (clab) at (34em,12.668em) {$a$};
\draw (c) -- (ab);
\draw (c) -- (a);
\draw (c) -- (b);
% b/c 
%\node (bbc) at (6em,14.8em) {$b$};
\node (bc) at (28em,19.6em) {$x$};
\node (aabc) at (28.7em,19.2em) {};
\node (cbc) at (34em,19.6em) {$y$};
\node (aacbc) at (33.3em,19.2em) {};
\node[tns] (cc) at (31em,17.668em) {};
\node (clab) at (31em,17.668em) {$a$};
\draw (cc) -- (a);
\draw (cc) -- (aabc);
\draw (cc) -- (aacbc);
% |- a/c
\end{tikzpicture}
\caption{Structural rewrites for associativity of mode $a$.}
\label{fig:assoc}
\end{center}
\end{figure}
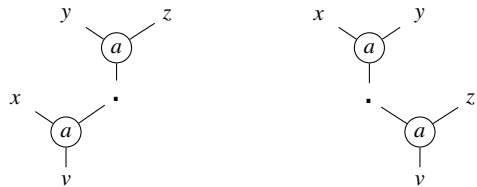

However, we have seen that this statement becomes derivable once we
add associativity of $a$ and it is easily verified to be a theorem of
the Lambek calculus. How can we add a modally controlled version of
associativity to the proof net calculus? We can add such a rule by
adding a rewrite from a tensor tree to another tensor tree with the
same set of leaves. The rewrite for associativity is shown in
Figure~\ref{fig:assoc}. To apply a structural rewrite, we replace the
tree on the left hand side of the arrow by the one on the right hand
side, reattaching the leaves and the root to the rest of the proof net.

Just like the structural rules, a structural rewrite always has the same leaves on both sides of the
arrow --- neither copying nor deletion is allowed\footnote{From the
  point of view of linear logic, we stay within the purely
  multiplicative fragment, which is simplest proof-theoretically.}, though we can
reorder the leaves in any way (the associativity rule doesn't reorder
the leaves). 

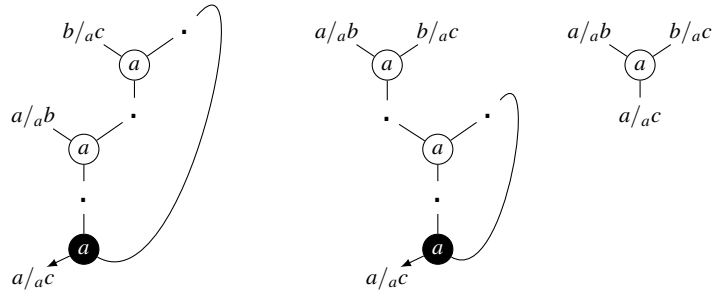
\begin{figure}
\begin{center}
\begin{tikzpicture}[scale=0.75]
% a/b 
%\node (labl) at (3em,7.5em) {$[\ldr E]$};
\node (ab) at (13em,9.8em) {$\apsnodei$};
\node (a) at (10em,14.6em) {$a/_a b$};
\node (aa) at (10.7em,14.2em) {};
\node (b) at (16em,14.75em) {$\apsnodei$};
\node[tns] (c) at (13em,12.668em) {};
\node (clab) at (13em,12.668em) {$a$};
\draw (c) -- (ab);
\draw (c) -- (aa);
\draw (c) -- (b);
% b/c 
%\node (bbc) at (6em,14.8em) {$b$};
\node (bc) at (13em,19.6em) {$b/_a c$};
\node (aabc) at (13.7em,19.2em) {};
\node (cbc) at (19em,19.75em) {$\apsnodei$};
\node[tns] (cc) at (16em,17.668em) {};
\node (clab) at (16em,17.668em) {$a$};
\draw (cc) -- (b);
\draw (cc) -- (aabc);
\draw (cc) -- (cbc);
% |- a/c
\node (pa) at (10em,5.0em) {$a/_a c$};
\node (pat) at (10.3em,5.44em) {};
\node[par] (pc) at (13em,6.732em) {};
\node (pclab) at (13em,6.732em) {\textcolor{white}{$a$}};
%\node (pb) at (21em,5.0em) {$c$};
%\node (pd) at (3em,9.8em) {$a$};
%\draw (pc) -- (pb);
\draw (pc) -- (ab);
\path[>=latex,->]  (pc) edge (pat);
\draw (cbc) to [out=50,in=330] (pc);
%
% a/b 
%\node (labl) at (3em,7.5em) {$[\ldr E]$};
\node (ab) at (34em,9.8em) {$\apsnodei$};
\node (a) at (31em,14.6em) {$\apsnodei$};
%\node (a) at (31em,14.6em) {$a/_a b$};
%\node (aa) at (31.7em,14.2em) {};
\node (b) at (37em,14.75em) {$\apsnodei$};
\node[tns] (c) at (34em,12.668em) {};
\node (clab) at (34em,12.668em) {$a$};
\draw (c) -- (ab);
\draw (c) -- (a);
\draw (c) -- (b);
% b/c 
%\node (bbc) at (6em,14.8em) {$b$};
\node (bc) at (28em,19.6em) {$a/_a b$};
\node (aabc) at (28.7em,19.2em) {};
\node (cbc) at (34em,19.6em) {$b/_a c$};
\node (aacbc) at (33.3em,19.2em) {};
\node[tns] (cc) at (31em,17.668em) {};
\node (clab) at (31em,17.668em) {$a$};
\draw (cc) -- (a);
\draw (cc) -- (aabc);
\draw (cc) -- (aacbc);
% |- a/c
\node (pa) at (31em,5.0em) {$a/_a c$};
\node (pat) at (31.3em,5.44em) {};
\node[par] (pc) at (34em,6.732em) {};
\node (pclab) at (34em,6.732em) {\textcolor{white}{$a$}};
%\node (pb) at (21em,5.0em) {$c$};
%\node (pd) at (3em,9.8em) {$a$};
%\draw (pc) -- (pb);
\draw (pc) -- (ab);
\path[>=latex,->]  (pc) edge (pat);
\draw (b) to [out=50,in=330] (pc);
%%%
\node (bc) at (43em,19.6em) {$a/_a b$};
\node (aabc) at (43.7em,19.2em) {};
\node (cbc) at (49em,19.6em) {$b/_a c$};
\node (aacbc) at (48.3em,19.2em) {};
\node[tns] (cc) at (46em,17.668em) {};
\node (clab) at (46em,17.668em) {$a$};
\node (a) at (46em,14.6em) {$a/_a c$};
\draw (cc) -- (a);
\draw (cc) -- (aabc);
\draw (cc) -- (aacbc);
\end{tikzpicture}
\caption{Structural rewrite and contraction for the abstract proof
  structure of Figure~\ref{fig:aps}, showing this is a proof net for $a/_a b
  \circ_a b/_a c \vdash a/_a c$}
\label{fig:red}
\end{center}
\end{figure}

Figure~\ref{fig:red} shows how the contractions and the structural
rewrites work together to derive $a/_a b, b/_a c \vdash a/_a c$. 

% TODO: a dot output somewhere

We start with a structural rewrite, which rebrackets the pair of
tensor links. The two hypotheses are now the premisses of the same
link, and this also produces a contractible structure for the $/I$
link. Hence, we have shown the proof structure to be a proof net.

\begin{wrapfigure}{r}{.5\textwidth}
\includegraphics[width=5cm]{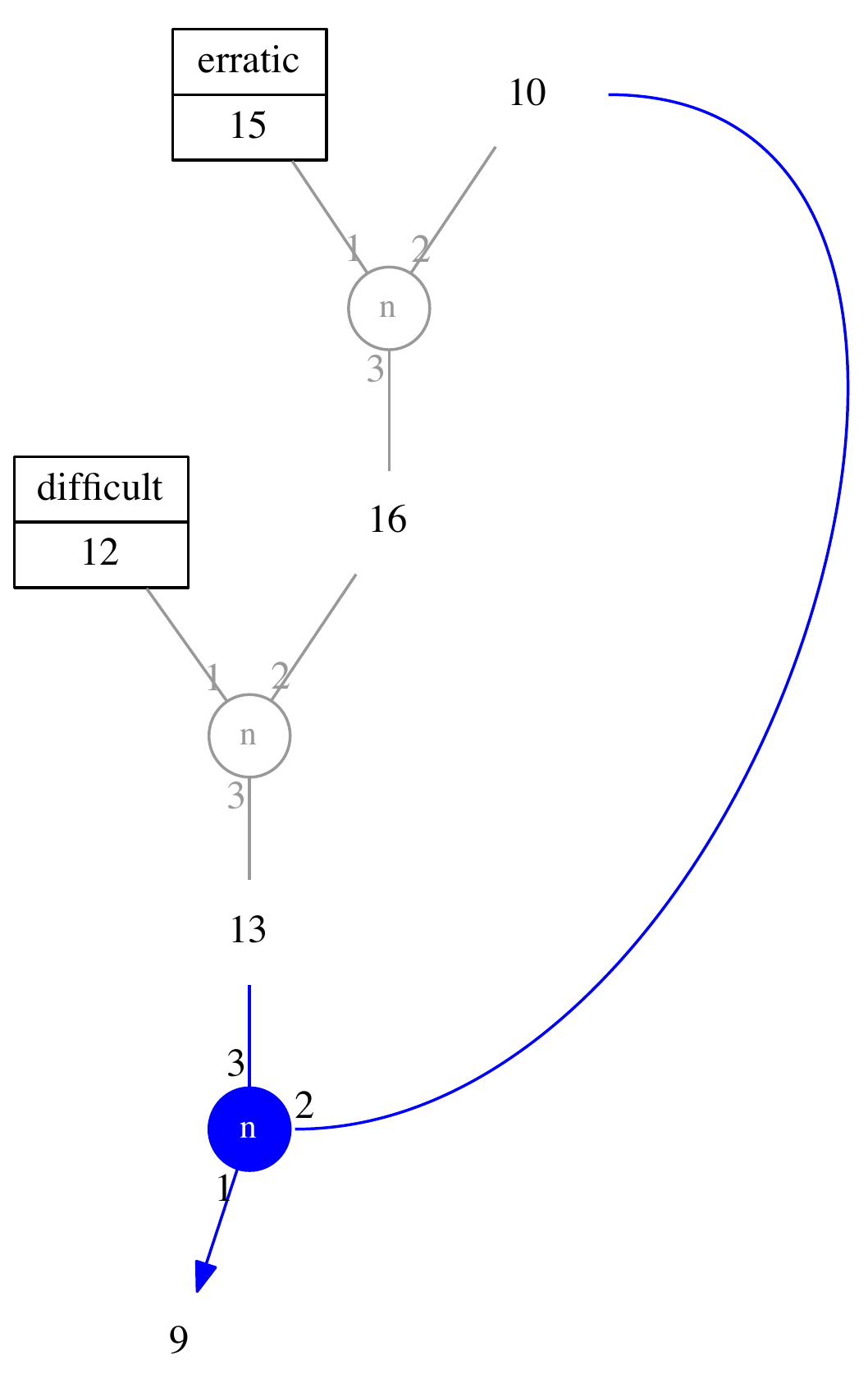}
\caption{Interactive Grail output}
\label{fig:grail}
\end{wrapfigure}

In the Grail theorem prover, the representation of abstract proof
structures looks as shown in Figure~\ref{fig:grail} (this is an automatically produced subgraph close
to the graph on the left of Figure~\ref{fig:red}, though with a
non-associative mode $n$ and therefore not derivable). This graph is used
during user interaction. The graphs are drawn using GraphViz, an
external graph drawing program which does not guarantee respecting our
desires for left, right and top/bottom, so
tentacles are labeled 1, 2 and 3 (for left, right and top/bottom
respectively) to allow us to make these distinctions regardless of the
visual representation. Vertices are given unique identifiers for user
interaction, for example to allow specifying which pair of atoms
should be identified or which par link should be contracted.

Although the structural rules give the grammar writer a great deal of
flexibility, such flexibility complicates proof search. As discussed
at the beginning of Section~\ref{sec:parse}, theorem
proving using proof nets is a four step process, which in the current
situation looks as follows: 1) lexical lookup, 2)
unfolding, 3) identification of atoms, 4) graph rewriting. In the
current case, both the graph rewriting and the identification of atoms
are complicated\footnote{Lexical ambiguity is a major problem for
  automatically extracted wide-coverage grammars as well, though
  standard statistical methods can help alleviate this problem
  \citep{moot10grail}.} and since we can interleave the atom
connections and the graph rewriting it is not a priori clear which
strategy is optimal for which set of structural rules. The current
implementation does graph rewriting only once all atoms have been
connected. 

The Grail theorem prover implements some strategies for early failure. Since all
proofs in multimodal type-logical grammars are a subset of the proofs
in multiplicative linear logic, we can reject (partial) proof
structures which are invalid in multiplicative linear logic, a
condition which is both powerful and easy to check.

As a compromise between efficiency and flexibility, Grail allows the
grammar writer to specify a first-order approximation of her
structural rules. Unlike the test for validity in multiplicative
linear logic which is valid for any set of structural rules,
specifying such a first-order approximation is valid only when there
is a guarantee that all derivable sequents in the multimodal grammar
are a subset of their approximations derivable in first-order linear
logic. Errors made here can be rather subtle and hard to detect. It is
recommended to use such methods to improve parsing speed only when a
grammar has been sufficiently tested and where it is possible to
verify whether no valid readings are excluded, or, ideally, to prove
that the subset relation holds between the multimodal logic and its
first-order approximation.

The next section will discuss first-order proof nets in their own
right. Though these proof nets have been used as an underlying
mechanism in Grail for a long time, we have seen in Section~\ref{sec:mill}
that many modern type-logical grammars are formulated in a way which
permits a direct implementation without an explicit set of structural rules.

As to the proof search strategy used by Grail, it is an instance of
the ``dancing links'' algorithm \citep{knuth00dancing}: when connecting
atomic formulas, we always link a formula which has the least
possibilities and we rewrite the abstract proof structures only once a
fully linked proof structure has been produced. Though the parser is
not extremely fast, evaluation both on randomly generated statements
and on multimodal statements extracted from corpora show that the
resulting algorithm performs more than well enough \citep{moot08filterr}.

\subsection{First-order proof nets}
\label{sec:millpn}

Proof nets for first-order linear logic \citep{quant} are a simple extension of the
proof nets for standard, multiplicative linear logic
\citep{multiplicatives}. Compared to the multimodal proof nets of the
previous section, all logical links have the main formula of the link
as their conclusion but there is now a notion of \emph{polarity},
corresponding to whether or not the formula occurs on the left hand
side of the turnstile (negative polarity) or on the right hand side
(positive polarity).

We unfold a sequent $A_1,\ldots,A_n \vdash C$ by using the negative
unfolding for each of the $A_i$ and the positive unfolding for
$C$. The links for first-order proof nets are shown in Table~\ref{tab:links}.

\begin{table}

\begin{center}
\begin{tikzpicture}
\node (anx) at (0em,0em) {$\overset{-}{A}$};
\node (anp) at (6em,0em) {$\overset{+}{A}$};
\draw (anx) -- (0em,2em) -- (6em,2em) -- (anp);
\node (cnx) at (12em,0em) {$\overset{-}{A}$};
\node (cnp) at (18em,0em) {$\overset{+}{A}$};
\draw (cnx) -- (12em,-2em) -- (18em,-2em) -- (cnp);
\end{tikzpicture}
\end{center}
\vspace{\baselineskip}
\begin{center}
\begin{tikzpicture}
\node (forallnc) {$\overset{-}{\forall x. A}$};
\node (forallnp) [above=2em of forallnc] {$\overset{-}{A[x:=t]}$};
\draw (forallnc) -- (forallnp);
\node (forallpc) [right=7em of forallnc] {$\overset{+}{\forall x. A}$};
\node (forallpp) [above=2em of forallpc] {$\overset{+}{A}$};
\draw[<-,semithick,dotted] (forallpc) -- (forallpp) node [midway] {$x\
  \ \ \ $};
%\draw [dotted] (forallpc) -- (forallpp);
%
\node (existsnc) [right=7em of forallpc] {$\overset{-}{\exists x. A}$};
\node (existsnp) [above=2em of existsnc] {$\overset{-}{A}$};
\draw[<-,semithick,dotted] (existsnc) -- (existsnp) node [midway] {$x\
  \ \ \ $};
\node (otimesnc) [above=7em of existsnc] {$\overset{-}{A\otimes B}$};
\node (tmponl) [left=0.66em of otimesnc] {};
\node (aotimesnc) [above=2.5em of tmponl] {$\overset{-}{A}$};
\node (tmponr) [right=0.66em of otimesnc] {};
\node (botimesnc) [above=2.5em of tmponr] {$\overset{-}{B}$};
\begin{scope}
\begin{pgfinterruptboundingbox}
\path [clip] (otimesnc.center) circle (2.5ex) [reverseclip];
\end{pgfinterruptboundingbox}
\draw [dotted] (otimesnc.center) -- (botimesnc);
\draw [dotted] (otimesnc.center) -- (aotimesnc);
\end{scope}
\begin{scope}
\path [clip] (aotimesnc) -- (otimesnc.center) -- (botimesnc);
\draw (otimesnc.center) circle (2.5ex);
\end{scope}
\node (otimespc) [right=7em of otimesnc] {$\overset{+}{A\otimes B}$};
\node (tmpopl) [left=0.66em of otimespc] {};
\node (aotimespc) [above=2.5em of tmpopl] {$\overset{+}{A}$};
\node (tmpopr) [right=0.66em of otimespc] {};
\node (botimespc) [above=2.5em of tmpopr] {$\overset{+}{B}$};
\draw (otimespc) -- (aotimespc);
\draw (otimespc) -- (botimespc);
\node (existspc) [below=7em of otimespc] {$\overset{+}{\exists x. A}$};
\node (existspp) [above=2em of existspc] {$\overset{+}{A[x:=t]}$};
\draw  (existspc) -- (existspp);
\node (lollinc) [above=7em of forallnc] {$\overset{-}{A\multimap B}$};
\node (tmplnl) [left=0.66em of lollinc] {};
\node (alollin) [above=2.5em of tmplnl] {$\overset{+}{A}$};
\draw (lollinc) -- (alollin);
\node (tmplnr) [right=0.66em of lollinc] {};
\node (blollin) [above=2.5em of tmplnr] {$\overset{-}{B}$};
\draw (lollinc) -- (blollin);
\node (lollipc) [above=7em of forallpc] {$\overset{+}{A\multimap B}$};
\node (tmplpl) [left=0.66em of lollipc] {};
\node (alollip) [above=2.5em of tmplpl] {$\overset{-}{A}$};
\node (tmplpr) [right=0.66em of lollipc] {};
\node (blollip) [above=2.5em of tmplpr] {$\overset{+}{B}$};
\begin{scope}
\begin{pgfinterruptboundingbox}
\path [clip] (lollipc.center) circle (2.5ex) [reverseclip];
\end{pgfinterruptboundingbox}
\draw [dotted] (lollipc.center) -- (blollip);
\draw [dotted] (lollipc.center) -- (alollip);
\end{scope}
\begin{scope}
\path [clip] (alollip) -- (lollipc.center) -- (blollip);
\draw (lollipc.center) circle (2.5ex);
\end{scope}
\end{tikzpicture}
\end{center}

\caption{Logical links for MILL1 proof structures}
\label{tab:links}
\end{table}

Contrary to multimodal proof nets, where a tensor link was drawn
with an open central node and a par link with a filled central node,
here par links are drawn as a connected pair of dotted lines and
tensor links as a pair of solid
lines. 

As before, premisses are drawn above the link and conclusions are
drawn below it. With the exception of the cut and axiom links, the order of
the premisses and the conclusions is important. We assume without loss
of generality that every quantifier link uses a distinct eigenvariable.

A set of formula occurrences connected by links is a proof
structure if every formula is at most once the premiss of a link and
if every formula is exactly once the conclusion of a link. Those
formulas which are not the premiss of any link are the conclusions of
the proof structure --- note the difference with multimodal proof
nets: a proof structure has conclusions but no hypotheses and,
as a consequence, each formula in the proof net must be the conclusion
of exactly one (instead of at most one) link.

For polarised proof nets, unfolding the formulas according to the
links of Table~\ref{tab:links} no longer produces a proof structure, since the atomic formulas after unfolding are not the conclusions of
any link. Such ``partial proof structures'' are called a modules.
To turn a module into a proof structure, we connect atomic
formulas of opposite polarity by axiom links until we obtain a
complete matching of the atomic formulas, that is until every atomic
formula is the conclusion of an axiom link.

The negative $\forall$ and the positive $\exists$ link, are defined
using substitution of an arbitrary term $t$ for the eigenvariable of
the link. In actual proof search, we use unification of these
variables when the axiom links are performed. 

As usual, not all proof structures are proof nets. However, since the
logical rules for the quantifiers make essential use of the notion of
``free occurrence of a variable'', this should be reflected in out
correctness condition. \citeasnoun{quant} uses a notion of switching
for proof structures which extends the switchings of
\citeasnoun{multiplicatives}.

A \emph{switching} is, for each of the binary par links a choice of
its left or right premiss and for each of the unary par links with
eigenvariable $x$ a choice of one of the formulas in the structure
with a free occurrence of $x$ \emph{or} of the premiss of the rule.

Given a switching, a \emph{correction graph} replaces a binary par link
by a connection from the conclusion of the link to the premiss chosen
by the switching, and it replace a unary par link by a link from the
conclusion to the formula chosen by the switching.

Finally, a proof structure is a \emph{proof net} when all its
correction graphs are both acyclic and connected \citep{quant}.

\begin{figure}
\begin{center}
\begin{tikzpicture}
\node (lollinc) {$\overset{-}{a\multimap \exists x. b(x)}$};
\node (tmplnl) [left=0.0em of lollinc] {};
\node (alollin) [above=2.5em of tmplnl] {$\overset{+}{a}$};
\draw (lollinc) -- (alollin);
\node (tmplnr) [right=0.0em of lollinc] {};
\node (blollin) [above=2.5em of tmplnr] {$\overset{-}{\exists x. b(x)}$};
\draw (lollinc) -- (blollin);
%
%
%\node (existsnc) [right=7em of forallpc] {$\overset{-}{\exists x. A}$};
\node (existsnp) [above=2em of blollin] {$\overset{-}{b(x)}$};
\draw[<-,semithick,dotted] (blollin) -- (existsnp) node [midway] {$x\
  \ \ \ $};
\node (existspc) [right=9em of lollinc] {$\overset{+}{\exists
    y. [a\multimap b(y)]}$};
\node (existspp) [above=2em of existspc] {$\overset{+}{a \multimap b(x)}$};
\draw  (existspc) -- (existspp);
%
%\node (existspp) [above=7em of forallpc] {$\overset{+}{A\multimap B}$};
\node (tmplpl) [left=0.66em of existspp] {};
\node (alollip) [above=2.5em of tmplpl] {$\overset{-}{a}$};
\node (tmplpr) [right=0.66em of existspp] {};
\node (blollip) [above=2.5em of tmplpr] {$\overset{+}{b(x)}$};
% b axiom
\coordinate[above=1em of existsnp] (y2);
\coordinate[above=1em of blollip] (y1);
\draw (blollip) -- (y1) -- (y2) -- (existsnp);
% a axiom
\coordinate[above=2.5em of alollip] (y2);
\coordinate[above=7em of alollin] (y1);
\draw (alollin) -- (y1) -- (y2) -- (alollip);
% par connection
\begin{scope}
\begin{pgfinterruptboundingbox}
\path [clip] (existspp.center) circle (2.5ex) [reverseclip];
\end{pgfinterruptboundingbox}
\draw [dotted] (existspp.center) -- (blollip);
\draw [dotted] (existspp.center) -- (alollip);
\end{scope}
\begin{scope}
\path [clip] (alollip) -- (existspp.center) -- (blollip);
\draw (existspp.center) circle (2.5ex);
\end{scope}
\end{tikzpicture}
\end{center}
\caption{Proof structure for $a\multimap \exists x.b(x) \vdash \exists
  y. [a\multimap b(y)]$.}
\label{fig:lpn}
\end{figure}
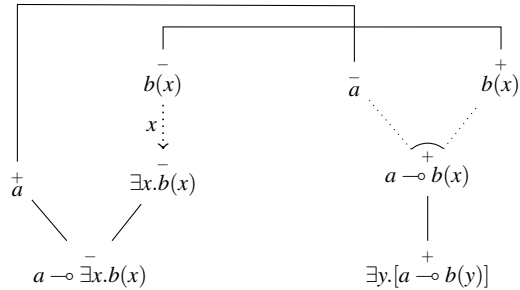

As an example, look at the proof structure of $a\multimap \exists x.b(x) \vdash \exists
  y. [a\multimap b(y)]$ shown in Figure~\ref{fig:lpn}. This statement
  is \emph{not} derivable in first-order linear logic (nor in
  intuitionistic logic). Consider therefore the switching connecting
  the binary par link to its left premiss $a$ and the link for $x$ to
  the formula $a\multimap b(x)$ (it has a free occurrence of $x$, so
  this like is a valid switching).

\begin{figure}
\begin{center}
\begin{tikzpicture}
\node (lollinc) {$\overset{-}{a\multimap \exists x. b(x)}$};
\node (tmplnl) [left=0.0em of lollinc] {};
\node (alollin) [above=2.5em of tmplnl] {$\overset{+}{a}$};
\draw[thick] (lollinc) -- (alollin);
\node (tmplnr) [right=0.0em of lollinc] {};
\node (blollin) [above=2.5em of tmplnr] {$\overset{-}{\exists x. b(x)}$};
\draw[thick] (lollinc) -- (blollin);
\node (existsnp) [above=2em of blollin] {$\overset{-}{b(x)}$};
\node (existspc) [right=9em of lollinc] {$\overset{+}{\exists
    y. [a\multimap b(y)]}$};
\node (existspp) [above=2em of existspc] {$\overset{+}{a \multimap b(x)}$};
\draw  (existspc) -- (existspp);
%
%\node (existspp) [above=7em of forallpc] {$\overset{+}{A\multimap B}$};
\node (tmplpl) [left=0.66em of existspp] {};
\node (alollip) [above=2.5em of tmplpl] {$\overset{-}{a}$};
\node (tmplpr) [right=0.66em of existspp] {};
\node (blollip) [above=2.5em of tmplpr] {$\overset{+}{b(x)}$};
% b axiom
\coordinate[above=1em of existsnp] (y2);
\coordinate[above=1em of blollip] (y1);
\draw (blollip) -- (y1) -- (y2) -- (existsnp);
% a axiom
\coordinate[above=2.5em of alollip] (y2);
\coordinate[above=7em of alollin] (y1);
\draw[thick] (alollin) -- (y1) -- (y2) -- (alollip);
\draw[thick] (blollin) -- (existspp);
\draw[thick] (existspp) -- (alollip);
\end{tikzpicture}
\caption{Correction graph for the proof structure of Figure~\ref{fig:lpn}
  with the cycle indicated, showing $a\multimap \exists x.b(x) \vdash \exists
  y. [a\multimap b(y)]$ is underivable}
\label{fig:cycle}
\end{center}
\end{figure}
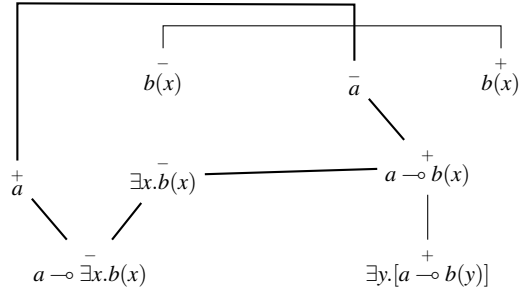

This switching produces the correction graph shown in
Figure~\ref{fig:cycle}. It contains a cycle, drawn with bold edges,
and is therefore not a proof structure (in addition, the $b$ axiom is disconnected from the rest of the
structure, giving a second reason for rejecting the proof structure).

\paragraph{Contractions} Though switching conditions for proof nets
are simple and elegant, they don't lend themselves to naive
application: already for the example proof structure of
Figure~\ref{fig:lpn} there are six possible switchings to consider
and, as the reader can verify, only the switching shown in
Figure~\ref{fig:cycle} is cyclic (and disconnected). In general, it
is often the case that all switchings but one are acyclic and
connected, as it is here.

Though there are efficient ways of testing acyclicity and
connectedness for multiplicative proof nets \citep{pnlinear,murong} and
it seems these can be adapted to the first-order case (though some
care needs to be taken when we allow complex terms), the theorem
prover for first-order linear logic uses a extension of the contraction
criterion of \citeasnoun{reductions}.

Given a proof structure we erase all formulas from the vertices and
keep only a set of the free variables at this vertex. We then use the
contractions of Table~\ref{fig:controne} to contract the edges of the
graph. The resulting vertex of each contraction has the union of the
free variables of the two vertices of the redex (we remove the
eigenvariable $x$ of a $\forall$ contraction, `` $\Rightarrow_u$''). 
% (the multimodal proof nets of Section~\ref{sec:mmpn}
%nets use a different variant of this contraction criterion).
A proof structure is a proof net iff it contracts to a single vertex
using the contractions of Table~\ref{fig:controne}.

\begin{table}
\begin{center}
\begin{tikzpicture}
\node (x) at (0em,0em) {$v_i$};
\node (y) at (0em,4em) {$v_j$};
\draw [semithick,dotted] plot [smooth, tension=1] coordinates {(-0.4em,0.5em) (-1em,2em) (-0.4em,3.5em)};
\draw [semithick,dotted] plot [smooth, tension=1] coordinates {(0.4em,0.5em) (1em,2em) (0.4em,3.5em)};
\draw plot [smooth,tension=1] coordinates {(-0.4em,0.5em) (0em,0.7em) (0.4em,0.5em)};
%
%\node (x2) at (4em,0em) {$v_i$};
\node (x2) at (4em,2em) {$v_i$};
%\node (y2) at (4em,4em) {$v_j$};
%\draw (x2) -- (y2);
\node (a1) at (2.5em,2em) {$\Rightarrow_{\textit{p}}$};
\node (x3) at (10em,0em) {$v_i$};
\node (y3) at (10em,4em) {$v_j$};
\draw[<-,semithick,dotted] (x3) -- (y3);
\node (xv) at (9.45em,2.15em) {$x$};
%\node (x4) at (14em,0em) {$v_i$};
%\node (y4) at (14em,4em) {$v_j$};
%\draw (x4) -- (y4);
\node (x4) at (14em,2em) {$v_i$};
\node (a2) at (12em,2em) {$\Rightarrow_{\textit{u}}$};
\node (x3) at (20em,0em) {$v_i$};
\node (y3) at (20em,4em) {$v_j$};
\draw (x3) -- (y3);
\node (x4) at (24em,2em) {$v_i$};
\node (a3) at (22em,2em) {$\Rightarrow_{\textit{c}}$};
\end{tikzpicture}
\end{center}
\caption{Contractions for first-order linear logic. Conditions: $v_i \neq v_j$
  and, for the $u$ contraction, all free occurrences of $x$ are at $v_j$.}
\label{fig:controne}
\end{table}

\begin{figure}
\begin{center}
\begin{tikzpicture}
\node (arrow) at (5.75cm,2.25cm) {$\Rightarrow_*$};
\node  (ga) at (9.0cm,3cm) {$\{ x\}$};
\node  (gb) at (7.0cm,3cm) {$\emptyset$};
\node (g2) at (8cm,1.5cm) {$\{ x \}$};
\draw[->,semithick,dotted] (ga) -- (gb);
\draw[semithick,dotted] (8.3cm,1.9cm) -- (ga);
\draw[semithick,dotted] (7.7cm,1.9cm) -- (gb);
\draw plot [smooth,tension=1] coordinates {(7.7cm,1.9cm) (8cm,2.05cm)
  (8.3cm,1.9cm)};
\node (lx) at (8.0cm,3.25cm) {$x$};
\node (lollinc) {$\emptyset$};
\node (tmplnl) [left=0.8em of lollinc] {};
\node (alollin) [above=2.5em of tmplnl] {$\emptyset$};
\draw (lollinc) -- (alollin);
\node (tmplnr) [right=0.8em of lollinc] {};
\node (blollin) [above=2.5em of tmplnr] {$\emptyset$};
\draw (lollinc) -- (blollin);
\node (existsnp) [above=2em of blollin] {$\{ x \}$};
\draw[<-,semithick,dotted] (blollin) -- (existsnp) node [midway] {$x\
  \ \ \ $};
\node (existspc) [right=9em of lollinc] {$\emptyset$};
\node (existspp) [above=2em of existspc] {$\{ x \}$};
\draw  (existspc) -- (existspp);
%
%\node (existspp) [above=7em of forallpc] {$\overset{+}{A\multimap B}$};
\node (tmplpl) [left=0.66em of existspp] {};
\node (alollip) [above=2.5em of tmplpl] {$\emptyset$};
\node (tmplpr) [right=0.66em of existspp] {};
\node (blollip) [above=2.5em of tmplpr] {$\{ x \}$};
% b axiom
\coordinate[above=1em of existsnp] (y2);
\coordinate[above=1em of blollip] (y1);
\draw (blollip) -- (y1) -- (y2) -- (existsnp);
% a axiom
\coordinate[above=2.5em of alollip] (y2);
\coordinate[above=6.0em of alollin] (y1);
\draw (alollin) -- (y1) -- (y2) -- (alollip);
% par connection
\begin{scope}
\begin{pgfinterruptboundingbox}
\path [clip] (existspp.center) circle (2.5ex) [reverseclip];
\end{pgfinterruptboundingbox}
\draw [dotted] (existspp.center) -- (blollip);
\draw [dotted] (existspp.center) -- (alollip);
\end{scope}
\begin{scope}
\path [clip] (alollip) -- (existspp.center) -- (blollip);
\draw (existspp.center) circle (2.5ex);
\end{scope}
\end{tikzpicture}
\end{center}
\caption{Contractions for the underivable $a\multimap \exists x.b(x) \vdash \exists
  y. [a\multimap b(y)]$.}
\label{fig:lpncontr}
\end{figure}
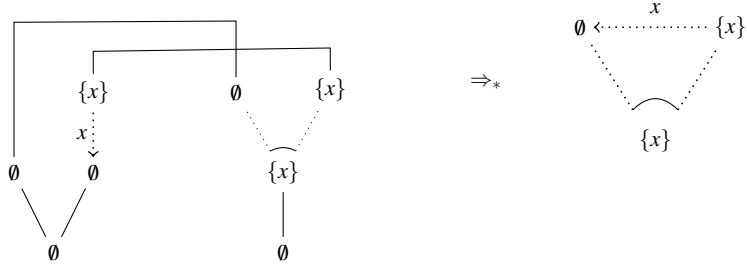

To give an example of the contractions, Figure~\ref{fig:lpncontr} shows
the contractions for the underivable proof structure of
Figure~\ref{fig:lpn}. The initial structure, which simply takes the
proof structure of
Figure~\ref{fig:lpn} and replaces the formulas by the corresponding
set of free variables, is shown on the left. Contracting the five
solid edges using the $c$ contraction produces the structure shown in
the figure on
the right. 

No further contractions apply: the two connected dotted links from the
binary par link do not end in the same vertex, so the par contraction
$p$ cannot apply. In addition, the universal contraction $u$ cannot
apply either, since it requires all vertices with its eigenvariable
$x$ to occur at the node from which the arrow is leaving and there is
another occurrence of $x$ at the bottom node of the structure. We have
therefore shown that this is not a proof net.

Since there are no structural rewrites, the contractions for
first-order linear logic are easier to apply than those for multimodal
type-logical grammars: it is rather easy to show confluence for the
contractions (the presence of structural rules, but also the unary
versions of the multimodal contractions, means confluence is not
guaranteed for multimodal proof nets). We already implicitly used
confluence when we argued that the proof structure in
Figure~\ref{fig:lpncontr} was not a proof net. The theorem prover uses a
maximally contracted representation of the proof structure to
represent the current state of proof search and this means less
overhead and more opportunities for early failure during proof search.

Like before, the theorem
proving uses four steps, which look as follows in the first-order case: 1) lexical lookup, 2)
unfolding, 3) axiom links with unification, 4) graph contraction.
Unlike the multimodal proof nets of the previous section, the graph
contractions are now confluent and can be performed efficiently
(the linear time solutions for the multiplicative case may be
adaptable, but a naive implementation already has an $O(n^2)$
worst-case performance). After lexical lookup, theorem proving for
first-order linear logic unfolds the formulas as before, but uses a
greedy contraction strategy. This maximally contracted partial proof
net constrains further axiom links: for example, a vertex containing a
free variable $x$ cannot be linked to the conclusion of the edge of
its eigenvariable (the vertex to which the arrow of the edge with
variable $x$ points) or to one of its descendants, since such a
structure would fail to satisfy the condition that the two vertices of
a $\forall$ link for the $u$ contraction of Figure~\ref{fig:controne}
are distinct. Another easily verified constraint is that two atomic
formulas can only be connected by an axiom link if these formulas
unify\footnote{As discussed in Section~\ref{sec:mmpn}, the multimodal
theorem prover allows the grammar writer to specify first-order
approximations of specific formulas. So underneath the surface of
Grail there is some first-order reasoning going on as well.}. Like for multimodal proof nets, the first-order linear logic
theorem prover chooses an axiom link for one of the atoms with the
fewest possibilities.

\subsection{Tools}

Table~\ref{tab:provers} lists the different theorem provers which are
available. Grail 0 \citep{moot15grail0} and Grail 3 \citep{grail3} use the multimodal proof net calculus
of Section~\ref{sec:mmpn}, whereas LinearOne \citep{moot15l1} uses the first-order
proof nets of Section~\ref{sec:millpn}.
GrailLight \citep{graillight} is a special-purpose chart parser, intended for use with an
automatically extracted French grammar for wide-coverage parsing and
semantics \citep{moot10grail,moot12spatio}. All provers are provided
under the GNU Lesser General Public License --- this means, notably,
there is no warranty, though I am committed to making all software as
useful as possible;  so contact me for any comments, feature requests or bug reports. All
theorem provers can be downloaded from
the author's GitHub site.

\medskip
\texttt{https://github.com/RichardMoot/}
\medskip

The columns of table Table~\ref{tab:provers}  indicate whether the theorem provers provide natural deduction
output, graph output (of the partial proof nets), whether there is an
interactive mode for proof search, whether the implementation is
complete and whether the grammar can specify its own set of structural rules;
``NA'' means the question doesn't apply to the given system
(GrailLight doesn't use a graphs to represent proofs and first-order
linear logic does not have a grammar-specific set of structural
rules). The table should help you select the most adequate tool for
your purposes.

LinearOne provides natural deduction output not
only for first-order linear logic, but also for the Displacement
calculus, hybrid type-logical grammars and lambda grammars. That is,
the grammar writer can write a grammar in any of these formalisms,
LinearOne will do proof search of the translation of this grammar in first-order linear logic and then
translate any resulting proofs back to the source language.

\begin{table}
\begin{tabular}{l|ccccc}
Prover & ND  & Graph & Interactive & Complete &
                                                                    User-defined SR
  \\ \hline
Grail 0 & + & -- & -- & + & + \\
Grail 3 & -- & + & + & + & + \\
GrailLight & + & NA & + & -- & -- \\
LinearOne & + & + & -- & + & NA \\
\end{tabular}
\caption{The different theorem provers}
\label{tab:provers}
\end{table}

The syntactic example proofs in this chapter have been automatically generated
using these tools and the corresponding grammars files, as well as
many other example grammars, are included in the repository.

%\section{Conclusions}
%
%I have given a high-level description of a number of theorem provers
%for modern type-logical grammars. 

\bibliographystyle{spbasic}
\bibliography{moot}

\begin{thebibliography}{47}
\providecommand{\natexlab}[1]{#1}
\providecommand{\url}[1]{{#1}}
\providecommand{\urlprefix}{URL }
\expandafter\ifx\csname urlstyle\endcsname\relax
  \providecommand{\doi}[1]{DOI~\discretionary{}{}{}#1}\else
  \providecommand{\doi}{DOI~\discretionary{}{}{}\begingroup
  \urlstyle{rm}\Url}\fi
\providecommand{\eprint}[2][]{\url{#2}}

\bibitem[{Ajdukiewicz(1935)}]{ajd}
Ajdukiewicz K (1935) Die syntaktische {K}onnexit{\" a}t. Studies in Philosophy
  1:1--27

\bibitem[{Asher(2011)}]{asher11web}
Asher N (2011) Lexical Meaning in Context: A Web of Words. Cambridge University
  Press

\bibitem[{Bar-Hillel(1953)}]{quasi}
Bar-Hillel Y (1953) A quasi-arithmetical notation for syntactic description.
  Language 29(1):47--58

\bibitem[{Barker and Shan(2014)}]{bs14cont}
Barker C, Shan C (2014) Continuations and Natural Language. Oxford Studies in
  Theoretical Linguistics, Oxford University Press

\bibitem[{Bassac et~al(2010)Bassac, Mery, and Retor\'{e}}]{bmr10tt}
Bassac C, Mery B, Retor\'{e} C (2010) Towards a type-theoretical account of
  lexical semantics. Journal of Logic, Language and Information 19(2):229--245,
  \doi{10.1007/s10849-009-9113-x},
  \urlprefix\url{http://dx.doi.org/10.1007/s10849-009-9113-x}

\bibitem[{van Benthem(1995)}]{benthem}
van Benthem J (1995) Language in Action: Categories, Lambdas and Dynamic Logic.
  MIT Press, Cambridge, Massachusetts

\bibitem[{Carpenter(1994)}]{carpenter1994natural}
Carpenter B (1994) A natural deduction theorem prover for type-theoretic
  categorial grammars. Tech. rep., Carnegie Mellon Laboratory for Computational
  Linguistics, Pittsburgh, Pennsylvania

\bibitem[{Chatzikyriakidis(2015)}]{stergios2015coq}
Chatzikyriakidis S (2015) Natural language reasoning using {C}oq: Interaction
  and automation. In: Proceedings of Traitement Automatique des Langues
  Naturelles ({TALN 2015})

\bibitem[{Danos(1990)}]{reductions}
Danos V (1990) La logique lin{\'e}aire appliqu{\'e}e {\`a} l'{\'e}tude de
  divers processus de normalisation (principalement du $\lambda$-calcul). PhD
  thesis, University of {Paris} {VII}

\bibitem[{Danos and Regnier(1989)}]{multiplicatives}
Danos V, Regnier L (1989) The structure of multiplicatives. Archive for
  Mathematical Logic 28:181--203

\bibitem[{Girard(1987)}]{Girard}
Girard JY (1987) Linear logic. Theoretical Computer Science 50:1--102

\bibitem[{Girard(1991)}]{quant}
Girard JY (1991) Quantifiers in linear logic {II}. In: Corsi G, Sambin G (eds)
  Nuovi problemi della logica e della filosofia della scienza, CLUEB, Bologna,
  Italy, vol~II, proceedings of the conference with the same name, Viareggio,
  Italy, January 1990

\bibitem[{Girard et~al(1995)Girard, Lafont, and Regnier}]{G95}
Girard JY, Lafont Y, Regnier L (eds)  (1995) Advances in Linear Logic. London
  Mathematical Society Lecture Notes, Cambridge University Press

\bibitem[{Guerrini(1999)}]{pnlinear}
Guerrini S (1999) Correctness of multiplicative proof nets is linear. In:
  Fourteenth Annual {IEEE} Symposium on Logic in Computer Science, {IEEE}
  Computer Science Society, pp 454--263

\bibitem[{Huijbregts(1984)}]{weakin}
Huijbregts R (1984) The weak inadequacy of context-free phrase structure
  grammars. In: de~Haan G, Trommelen M, Zonneveld W (eds) Van Periferie naar
  Kern, Foris, Dordrecht

\bibitem[{Knuth(2000)}]{knuth00dancing}
Knuth DE (2000) Dancing links. arXiv preprint cs/0011047

\bibitem[{Kubota and Levine(2012)}]{kl12gap}
Kubota Y, Levine R (2012) Gapping as like-category coordination. In: B\'{e}chet
  D, Dikovsky A (eds) Logical Aspects of Computational Linguistics, Springer,
  Nantes, Lecture Notes in Computer Science, vol 7351, pp 135--150

\bibitem[{Lambek(1958)}]{lambek}
Lambek J (1958) The mathematics of sentence structure. American Mathematical
  Monthly 65:154--170

\bibitem[{Luo(2012{\natexlab{a}})}]{luo12types}
Luo Z (2012{\natexlab{a}}) Common nouns as types. In: Logical aspects of
  computational linguistics ({LACL}’2012), Springer, Lecture Notes in
  Artificial Intelligence, vol 7351

\bibitem[{Luo(2012{\natexlab{b}})}]{luo12formal}
Luo Z (2012{\natexlab{b}}) Formal semantics in modern type theories with
  coercive subtyping. Linguistics and Philosophy 35(6):491--513

\bibitem[{Luo(2015)}]{luo15lambek}
Luo Z (2015) A {L}ambek calculus with dependent types. In: Types for Proofs and
  Programs ({TYPES} 2015), Tallinn

\bibitem[{Mery et~al(2013)Mery, Moot, and Retor\'{e}}]{mmr13plurals}
Mery B, Moot R, Retor\'{e} C (2013) Plurals: individuals and sets in a richly
  typed semantics. In: The Tenth International Workshop of Logic and
  Engineering of Natural Language Semantics 10 ({LENLS}10)

\bibitem[{Mineshima et~al(2015)Mineshima, Mart\'{\i}nez-G\'{o}mez, Miyao, and
  Bekki}]{koji2015coq}
Mineshima K, Mart\'{\i}nez-G\'{o}mez P, Miyao Y, Bekki D (2015) Higher-order
  logical inference with compositional semantics. In: Proceedings of Empirical
  Method for Natural Language Processing ({EMNLP} 2015)

\bibitem[{Montague(1970)}]{ug}
Montague R (1970) Universal grammar. Theoria 36(3):373--398

\bibitem[{Montague(1974)}]{montague}
Montague R (1974) The proper treatment of quantification in ordinary {English}.
  In: Thomason R (ed) Formal Philosophy. {Selected} Papers of Richard Montague,
  Yale University Press, New Haven

\bibitem[{Moortgat(2011)}]{Moo11}
Moortgat M (2011) Categorial type logics. In: van Benthem J, ter Meulen A (eds)
  Handbook of Logic and Language, North-Holland Elsevier, Amsterdam, chap~2, pp
  95--179

\bibitem[{Moortgat and Oehrle(1994)}]{mo94}
Moortgat M, Oehrle RT (1994) Adjacency, dependency and order. In: Proceedings
  9th {Amsterdam} Colloquium, pp 447--466

\bibitem[{Moot(2008)}]{moot08filterr}
Moot R (2008) Filtering axiom links for proof nets. Tech. rep., CNRS and
  Bordeaux University

\bibitem[{Moot(2010)}]{moot10grail}
Moot R (2010) Wide-coverage {French} syntax and semantics using {Grail}. In:
  Proceedings of Traitement Automatique des Langues Naturelles (TALN),
  Montreal, system Demo

\bibitem[{Moot(2012)}]{moot12spatio}
Moot R (2012) Wide-coverage semantics for spatio-temporal reasoning. Traitement
  Automatique des Languages 53(2):115--142

\bibitem[{Moot(2015{\natexlab{a}})}]{grail3}
Moot R (2015{\natexlab{a}}) Grail.
  \texttt{http://www.labri.fr/perso/moot/grail3.html}, mature and flexible
  parser for multimodal grammars

\bibitem[{Moot(2015{\natexlab{b}})}]{graillight}
Moot R (2015{\natexlab{b}}) Grail light.
  \texttt{https://github.com/RichardMoot/GrailLight}, fast, lightweight version
  of the {G}rail parser

\bibitem[{Moot(2015{\natexlab{c}})}]{moot15l1}
Moot R (2015{\natexlab{c}}) Linear one: A theorem prover for first-order linear
  logic. \texttt{https://github.com/RichardMoot/LinearOne}

\bibitem[{Moot and Piazza(2001)}]{mill1}
Moot R, Piazza M (2001) Linguistic applications of first order multiplicative
  linear logic. Journal of Logic, Language and Information 10(2):211--232

\bibitem[{Moot and Puite(2002)}]{mp}
Moot R, Puite Q (2002) Proof nets for the multimodal {Lambek} calculus. Studia
  Logica 71(3):415--442

\bibitem[{Moot and Retor\'{e}(2011)}]{mr11plurals}
Moot R, Retor\'{e} C (2011) Second order lambda calculus for meaning assembly:
  on the logical syntax of plurals. In: Computing Natural Reasoning (COCONAT),
  Tilburg

\bibitem[{Moot and Retor\'{e}(2012)}]{mr12lcg}
Moot R, Retor\'{e} C (2012) The Logic of Categorial Grammars: A Deductive
  Account of Natural Language Syntax and Semantics. No. 6850 in Lecture Notes
  in Artificial Intelligence, Springer

\bibitem[{Moot et~al(2015)Moot, Schrijen, Verhoog, and Moortgat}]{moot15grail0}
Moot R, Schrijen X, Verhoog GJ, Moortgat M (2015) Grail0: A theorem prover for
  multimodal categorial grammars.
  \texttt{https://github.com/RichardMoot/Grail0}

\bibitem[{Morrill et~al(2011)Morrill, Valent\'{\i}n, and
  Fadda}]{mvf11displacement}
Morrill G, Valent\'{\i}n O, Fadda M (2011) The displacement calculus. Journal
  of Logic, Language and Information 20(1):1--48

\bibitem[{Murawski and Ong(2000)}]{murong}
Murawski AS, Ong CHL (2000) Dominator trees and fast verification of proof
  nets. In: Logic in Computer Science, pp 181--191

\bibitem[{Oehrle(1994)}]{oehrle}
Oehrle RT (1994) Term-labeled categorial type systems. Linguistics \&
  Philosophy 17(6):633--678

\bibitem[{Pentus(1997)}]{pentus97}
Pentus M (1997) Product-free {Lambek} calculus and context-free grammars.
  Journal of Symbolic Logic 62:648--660

\bibitem[{Pogodalla and Pompigne(2012)}]{pp10acg}
Pogodalla S, Pompigne F (2012) Controlling extraction in abstract categorial
  grammars. In: de~Groote P, Nederhof MJ (eds) Proceedings of Formal Grammar
  2010--2011, Springer, LNCS, vol 7395, pp 162--177

\bibitem[{Pustejovsky(1995)}]{Pus95}
Pustejovsky J (1995) The generative lexicon. M.I.T. Press

\bibitem[{Ranta(1991)}]{ranta91icg}
Ranta A (1991) Intuitionistic categorial grammar. Linguistics and Philosophy
  14(2):203--239

\bibitem[{Shieber(1985)}]{shieber}
Shieber S (1985) Evidence against the context-freeness of natural language.
  Linguistics \& Philosophy 8:333--343

\bibitem[{Valent\'{\i}n(2014)}]{ov14}
Valent\'{\i}n O (2014) The hidden structural rules of the discontinuous
  {L}ambek calculus. In: Casadio C, Coecke B, Moortgat M, Scott P (eds)
  Categories and Types in Logic, Language, and Physics: Essays dedicated to
  {J}im {L}ambek on the Occasion of this 90th Birthday, no. 8222 in Lecture
  Notes in Artificial Intelligence, Springer, pp 402--420

\end{thebibliography}

\end{document}